%% file: PaperForReview.tex
\documentclass[10pt,twocolumn,letterpaper]{article}

\usepackage{titling}
\usepackage{cvpr}              %

\usepackage{graphicx}
\usepackage{amsmath}
\usepackage{amssymb} 
\usepackage{booktabs}

\usepackage[dvipsnames]{xcolor}

\usepackage{threeparttable}
\usepackage{tabularx}
\usepackage{array,multirow}

\usepackage{amsfonts,amsthm}
\usepackage{dsfont}

\DeclareMathOperator*{\argmin}{arg\,min}

\usepackage{subfiles}

\usepackage[pagebackref,breaklinks,colorlinks]{hyperref}

\usepackage[capitalize]{cleveref}
\crefname{section}{Sec.}{Secs.}
\Crefname{section}{Section}{Sections}
\Crefname{table}{Table}{Tables}
\crefname{table}{Tab.}{Tabs.}

\def\cvprPaperID{4780} %
\def\confName{CVPR}
\def\confYear{2023}

\newcolumntype{s}{>{\raggedleft\arraybackslash}X}

\def \dbscan {DBSCAN}
\def \dbscantrain {DBSCAN + init-train}
\def \dbscanst {DBSCAN + self-train}
\def \ppscore {PP score}
\def \ppscoretrain {PP score + init-train}
\def \modest {MODEST (1 traversal)}

\def \initlabel {point clustering pseudo-labels}

\def \ourmethod {OYSTER}
\def \ourmethodlong {\textbf{O}bject Discover\textbf{y} via \textbf{S}patio-\textbf{Te}mporal \textbf{R}efinement}

\begin{document}

\subfile{main_paper}

\newpage

\clearpage

\appendix

\section*{Appendix}

\subfile{main_supp}

\end{document}

%% file: main_paper.tex
%% LaTeX2e file `main_paper.tex'
%% generated by the `filecontents' environment
%% from source `PaperForReview' on 2023/06/16.
%%

\title{Towards Unsupervised Object Detection from LiDAR Point Clouds}

\makeatletter
\def\thanks#1{\protected@xdef\@thanks{\@thanks
        \protect\footnotetext{#1}}}
\makeatother
\author{
    \textbf{Lunjun Zhang \quad Anqi Joyce Yang \quad Yuwen Xiong \quad Sergio Casas} \\
    \textbf{Bin Yang${^\dagger}$ \quad Mengye Ren${^\dagger}$ \quad Raquel Urtasun}\\
    Waabi, University of Toronto \\
    \texttt {\{lzhang, jyang, yxiong, sergio, urtasun\}@waabi.ai}
    \thanks{$^\dagger$Work done at Waabi. Mengye is now at New York University.}
}

\maketitle

\begin{abstract}

In this paper, we study the problem of unsupervised object detection from 3D point clouds
in self-driving scenes.
We present a simple yet effective method that exploits (i) point clustering in near-range areas where the point clouds are dense, (ii) temporal consistency to filter out  noisy unsupervised detections, (iii) translation
equivariance of CNNs to extend the auto-labels to long range, and  (iv) self-supervision for improving on its own.
Our approach, \textbf{\ourmethod} (\ourmethodlong),
does not impose  constraints on data collection
(such as repeated traversals of the same location), is able to detect objects
in a zero-shot manner without supervised fine-tuning (even in sparse, distant regions), and continues to
self-improve given more rounds of iterative self-training.
To better measure model performance in self-driving scenarios, we propose a new planning-centric perception metric based
on  distance-to-collision.
We demonstrate that our unsupervised object detector significantly outperforms
unsupervised baselines on PandaSet and Argoverse 2 Sensor dataset, showing promise that self-supervision combined with object priors can enable object discovery in the wild. For more information, visit the project website: \url{https://waabi.ai/research/oyster}.

\end{abstract}
\vspace{-15pt}

\input{sec/intro}

\input{sec/related}

\input{sec/method}

\input{sec/exp}

\input{sec/conclusion}

\newpage

\section*{Acknowledgement}

We thank Arnaud Bonnet for infrastructure support, Yun Chen for valuable discussions along the way, and many others on the Waabi team. YX is partially supported by an NSERC scholarship and a Borealis AI fellowship. 

{\small
\bibliographystyle{ieee_fullname}
\bibliography{egbib}
}

%% file: sec/intro.tex
\section{Introduction}
\label{sec:intro}

\begin{figure}
  \centering
  \includegraphics[width=1.0\linewidth]{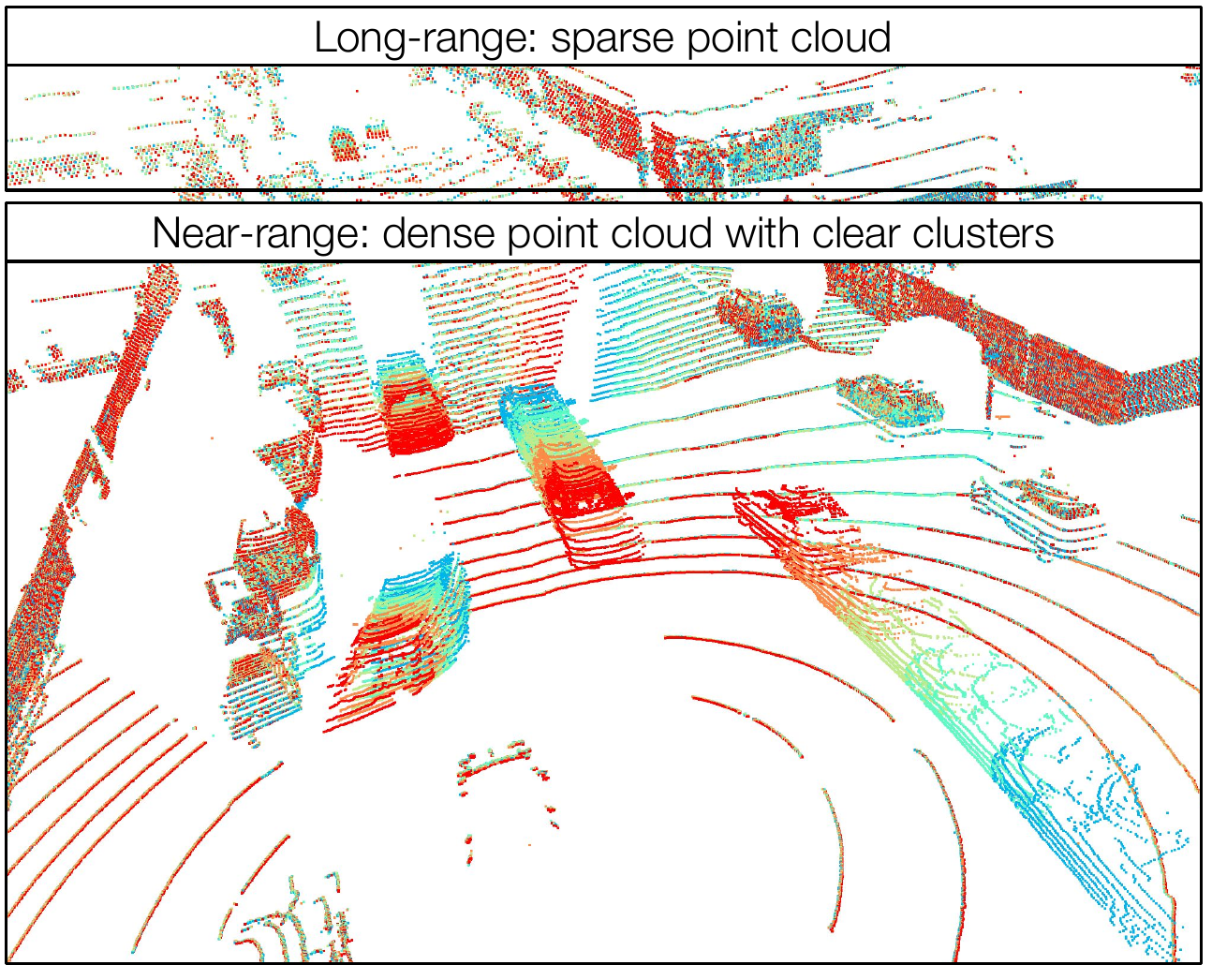}
  \vspace{-20pt}
  \caption{
    Visualization of a sequence of five 3D point clouds.
    At near range, the point clouds are very dense and we can clearly distinguish clusters, motivating us to use prior knowledge.
    At far range, the point clouds are quite sparse and the objects do not appear as obvious, leading us to explore zero-shot generalization.
  }
  \vspace{-20pt}
  \label{fig:hook}
\end{figure}

When a large set of annotations are available, supervised learning can solve the task of 3D object detection remarkably well thanks to the power of neural networks, as proven by many successful object detectors developed in the past decade \cite{faster-rcnn,mask-rcnn,fpn-net,center-net}.
However, since most existing data is unlabeled, human annotations are currently  the bottleneck for data-driven learning algorithms, as they require tedious manual effort that is very costly in practice. 
While there has been some effort on using weaker supervision to train object detectors \cite{min-entropy-weakly-supervised, convex-clustering, zigzag-learning}, it is worth noting that human and animal brains are able to perceive objects without explicit labels at all \cite{deep-learning-for-ai}. This naturally inspires us to ask whether we can design unsupervised learning algorithms that discover objects from raw streams of sensor data on their own.

Unsupervised object detection has long been studied in computer vision, albeit in various forms. For instance, numerous ways of unsupervised object proposals were considered as the first stage of an object detector \cite{rcnn}. These methods leverage a variety of  cues including colors and edge boundaries \cite{measure-objectness}, graph structures \cite{graph-based-segmentation}, and motion cues \cite{motion-cue-detect}. %
 While those methods are no longer popular in today's object detectors due to end-to-end supervised training, they offer important intuitions of \textit{what an object is}.

In recent years, unsupervised object detection has made a comeback under the name of \textit{object-centric models} \cite{kosiorek2018sequentialair, kosiorek2019scae, wang2021spair3d, burgess2019monet, engelcke2020genesis, lin2020space, locatello2020slot}. The essential idea is to train an auto-encoder with a structured decoder such that the network is forced to decompose the scene into a set of individual objects during reconstruction. Most of those models only show experiments on synthetic toy datasets, where the background lacks any fine-grained details, and the foreground objects have simplistic shapes and distinct colors. The reason why object-centric models have struggled to scale to realistic data is that their mechanism of object decomposition is based on careful balancing of model capacities between the foreground and background modules. Consequently, an increase in model capacity, which is needed for real-world data, breaks the brittle balance between modules and result in a failure in scene decomposition. Unsupervised object discovery in the wild remains an open challenge.

In this work, we study unsupervised object detection from point clouds in the context of self-driving vehicles (SDVs). 
This is a challenging task due to occlusion as well as  the sparsity of the observations particularly at range. We refer the reader to \cref{fig:hook} for an example. 
Despite its appeal, unsupervised object detection from LiDAR has received little attention. 
Recent work \cite{detect-mobile-objects} exploits repeated traversals of the same region to understand the persistence of a point over time and discover mobile objects from that information. However, the assumption of repeated traversals in acquiring point clouds restricts the applicability of the method. %
Instead, we study the most generic setting of unsupervised detection given any raw sequence of point clouds. %

Our method \textbf{\textit{\ourmethod}} (\textit{\ourmethodlong}) carefully combines key ideas from density-based spatial clustering, temporal consistency, equivariance and self-supervised learning in a unified framework that exploits their strengths while overcoming their shortcomings.
Firstly, we exploit point clustering to obtain initial pseudo-labels to bootstrap an object detector in the near range, where point density is high (see \cref{fig:hook}).
We then employ unsupervised tracking to filter out temporally inconsistent objects. 
Since point clustering does not work well in the long-range where observations are sparse, we exploit the translation equivariance of CNNs to train on high-quality, near-range pseudo-labels and zero-shot generalize to long range.
To bridge the density gap between short and long-range at training vs. inference, we propose a novel random LiDAR ray dropping strategy.
Finally, we design a self-improvement loop in which this bootstrapped model can self-train. 
At every round of self-improvement, we utilize the temporal consistency of objects to automatically refine the detections from the model at the previous iteration, and use these refined outputs as pseudo-labels for training.
Our experiments on Pandaset \cite{xiao2021pandaset} and Argoverse V2 Sensor \cite{wilson2021argoverse} demonstrate that \textit{\ourmethod} clearly outperforms other unsupervised methods, both under standard metrics based on intersection-over-union (IoU) and our proposed metric based on \textit{distance-to-collision} (DTC).
We hope that our work serves as a step towards building perception systems that automatically improve with more data and compute without being bottlenecked by human supervision.

%% file: sec/related.tex
\section{Related Work}
In this section, we briefly review the literature on three subfields that are closely related to unsupervised object detection: object-centric models, unsupervised object proposal (discovery), and open-set detection. 

\vspace{-10pt}
\paragraph{Object centric models:} A promising path towards unsupervised detection is broadly characterized as \textit{vision as inverse graphics}: learning deep generative models with structured decoder (or renderer) such that the encoder needs to decompose a scene into objects and parts and infer their orientations. In the deep learning era, this type of models are usually called object-centric models. In the single-image setting, many prior works have tried to use a mixture latent to encourage object discovery, such as AIR~\cite{eslami2016air}, Neural EM~\cite{greff2017neuralem,steenkiste2018relationalnem}, SCAE~\cite{kosiorek2019scae}, SPAIR~\cite{crawfordpineau2019spair}, MONet~\cite{burgess2019monet}, IODINE~\cite{greff2019iodine}, GENESIS~\cite{engelcke2020genesis}, SPACE~\cite{lin2020space}, Generative Neuro-Symbolic machine~\cite{jiang2020generative}, and Slot Attention~\cite{locatello2020slot}. Some effort has also been made in the setting of 3D view rendering, especially in light of recent progress in neural rendering models such as NeRF \cite{mildenhall2020nerf}: ObSuRF~\cite{stelzner2021obsurf} and Object Radiance Fields~\cite{yu2021objectradiance} try to combine Slot Attention with NeRF to discover objects given multi-view inputs. Certain object-centric models have attempted to take advantage of temporal and sequential information in videos, such as Sequential AIR~\cite{kosiorek2018sequentialair}, SPAIR with tracking~\cite{crawford2020spairtracking}, SCALOR~\cite{jiang2020scalor}, physics as inverse graphs~\cite{jaques2020physics}, SIMONe~\cite{kabra2021simone}, and Flow Capsule~\cite{sabour2021flowcapsule}. Those methods have only been shown to work on toy datasets.

\begin{figure*}
    \centering
    \vspace{-15pt}
    \includegraphics[width=0.9\linewidth]{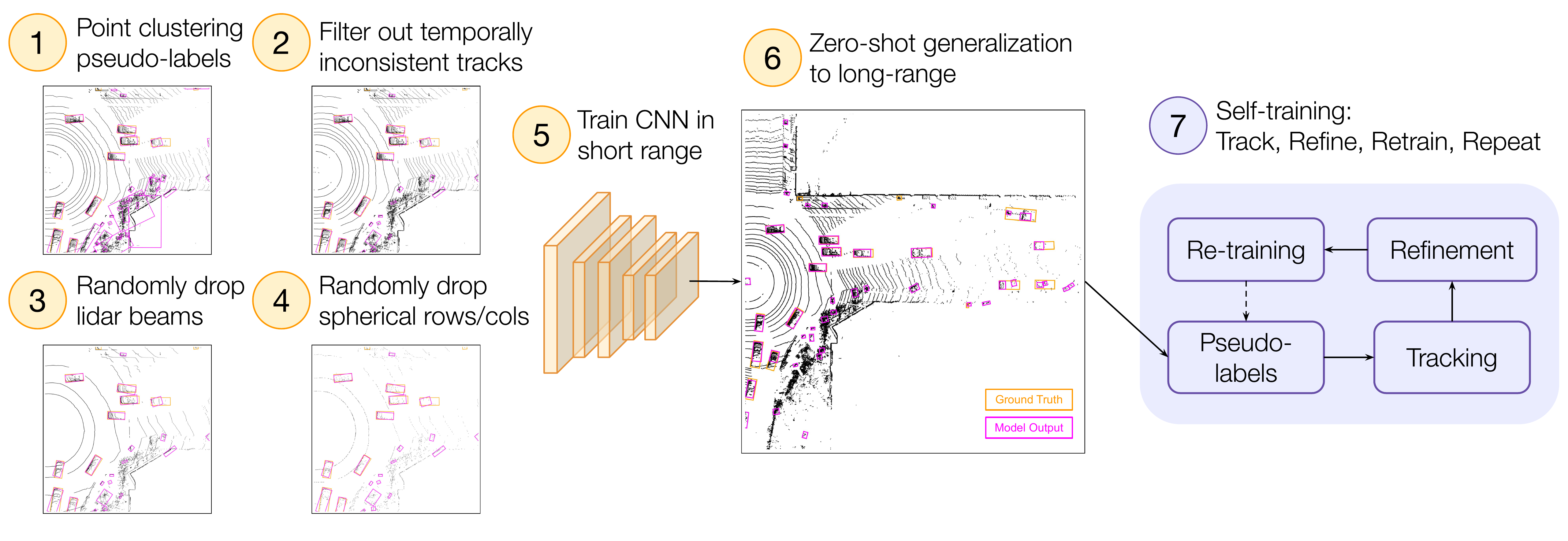}
    \vspace{-10pt}
    \caption{\textbf{Overview of \textit{\ourmethod}, our unsupervised detection method}. In the initial bootstrapping phase (step 1-6), we train a CNN on near-range point clustering results, and rely on translation equivalence of CNNs to produce pseudo-labels for full range. Our training applies random ray-dropping as data augmentation, and uses temporal-consistency based filtering on pseudo-labels. In the self-improvement phase (step 7), we propose a \textit{Track, Refine, Retrain, and Repeat} framework that teaches an unsupervised detector to iteratively self-improve.}
    \label{fig:overall-pipeline}
    \vspace{-10pt}
  \end{figure*}

\vspace{-10pt}
\paragraph{Unsupervised Object Discovery:} Many methods directly use various cues to discover objects without necessarily using a generative model. Prior works have explored Probabilistic Latent Semantic Analysis~\cite{sivic2005}, object co-segmentation~\cite{vicente2011coseg}, and a ranking based approach~\cite{vo2020discovery,vo2021discovery}. For 2D object segmentation and object proposals, existing approaches include parametric min cut~\cite{carreirasminchisescu2010}, selective search~\cite{uijlings2013}, and MCG~\cite{ponttuset2017}. When we have access to videos rather than just static images, motion information can be leveraged for object discovery, by utilizing motion segmentation~\cite{papazoglouferrari2013}, clustering the flow field~\cite{namdev2012}, directly segmenting moving object for detection~\cite{fragkiadaki2015}, or tracking and then segmenting~\cite{xiao2016}. In the case where we have 3D point clouds rather than an RGB image, prior techniques include point clustering \cite{douillard2011}, shape analysis~\cite{karpathy2013} and range-based analysis~\cite{bogoslavskyi2016}. Concurrently, \cite{najibi2022motion} leveraged scene flow estimation to generate seed labels for training an unsupervised LiDAR detector. Our method initially applies point clustering, but our core contribution is a learning-based method that can iteratively self-improve. In contrast to our approach, \cite{najibi2022motion} relies on unsupervised scene-flow to detect objects and thus is not able to detect static ones. Furthermore, it is limited to short ranges where point cloud alignment methods~\cite{mckayicp92} are reliable. \cite{detect-mobile-objects} does not need scene flow, but requires repeated traversals of the same location. Concurrent work \cite{4d-unsupervised-object-discovery} requires unsupervised scene flow and camera inputs for 3D instance segmentation. Our method requires neither scene flow, camera inputs, nor repeated traversals, so it can be directly applied to any LiDAR sequence.

\vspace{-10pt}
\paragraph{Open-Set Detection:} Another branch of methods deal with a weakly supervised setting where the system is given labels on certain known objects but is expected to discover the unknown stuff on its own~\cite{dhamija2020overlook}. For 2D object detection, prior art includes Bayesian approaches~\cite{pham2018bayesian} and dropout sampling for uncertainty estimation~\cite{miller2018dropout}. A new setting called Open-World Object Detection has been proposed as well~\cite{joseph2021openworld}, where the vision system sequentially receives new labeled instances on long-tailed classes~\cite{joseph2021openworld}. A Detection Transformer \cite{detr} for the Open-World setting has also been studied~\cite{gupta2021owdetr}. For 3D LiDAR object detection, prior methods include OSIS~\cite{wong2019osis} and open-set 3d-net~\cite{cen2021openset}; however, they do not study a completely unsupervised setting. 

%% file: sec/method.tex
\section{Method: \textit{\ourmethod}}

Our method has two phases of training: 
the \textbf{initial bootstrapping} phase, and the \textbf{self-improvement} phase.

The initial bootstrapping phase takes advantage of the fact that 
point clouds in the near range tend to be dense and have clear object clusters, 
so we can obtain reasonable near-range bounding box seed pseudo-labels via point clustering. 
Thanks to the translation equivariance property of convolutional nets, 
we find that a CNN detector trained on near-range labels can generalize 
to longer-range in a zero-shot manner with the help of data augmentations such as ray dropping that randomly sparsify the inputs. 

The self-improvement phase utilizes temporal consistency of object tracks 
as a self-supervision signal. Given noisy detections across time, 
we employ an unsupervised offline tracker to find object tracks of various lengths.
We discard short tracks, and refine long tracks. 
An object track should have the same object size across time, so our refinement process
uses track-level information to update pseudo-labels in long tracks. 
We train a new detector on the updated pseudo-labels, 
dump its outputs as new pseudo-labels, track, refine, and repeat. Fig.~\ref{fig:overall-pipeline} depicts the overall pipeline.

We describe the two training phases in more detail below.

\begin{figure*}
    \centering
    \vspace{-15pt}
    \includegraphics[width=1.0\linewidth]{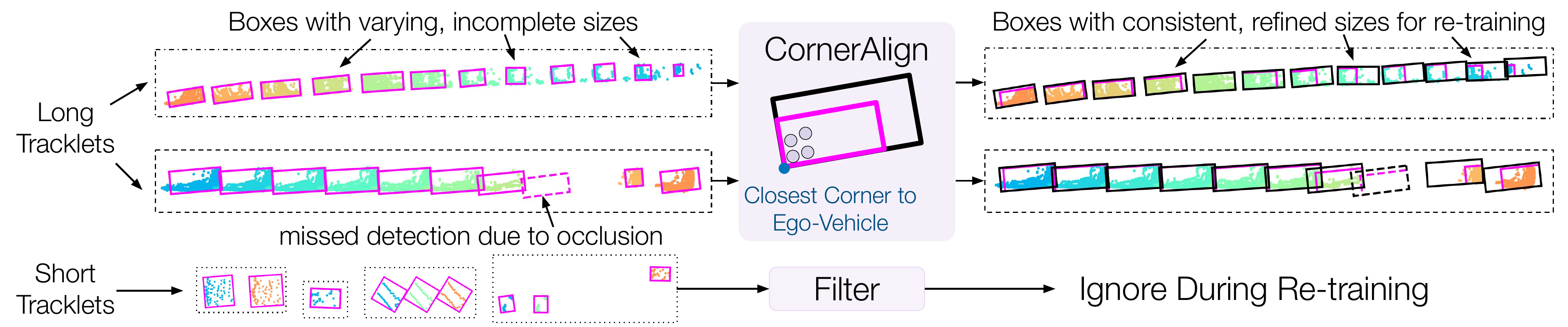}
    \vspace{-15pt}
    \caption{\textbf{Label Refinement Illustration}. With unsupervised tracking, we fill in missed detections and obtain per-object tracklets. For long tracklets, we leverage temporal observations to find a new, consistent object size and update the initial labels (in purple) with a corner-based alignment strategy~\cite{auto4d}. We additionally filter out short tracklet labels, which will not be used in the next round of self-training.}
    \label{fig:refinement}
    \vspace{-15pt}
  \end{figure*}

\subsection{Initial Bootstrapping via Point Clustering and Range Extension}

\label{sec:clustering-and-tracking}

Given a 3-dimensional LiDAR point cloud $p \in \mathbb{R}^{N\times 3}$
we obtain the initial seed pseudo-labels $\mathbf{B}^{(0)}$,
via ground removal, point clustering, and bounding box fitting:
\begin{align*}
    \mathbf{B}^{(0)} \leftarrow \text{FitBboxes}( \text{Clustering}(\text{RemoveGround}(p)) )
\end{align*}
where we employ the 2D Bird-Eye-View (BEV) detection representation~\cite{pointpillars,yang2018pixor} and each BEV bounding box $\mathbf{b} = (x,y,l,w,\theta) \in \mathbf{B}^{(0)}$ consists of the centroid position $(x,y)$, the length and width $(l,w)$, and the heading $\theta$.
We perform this for each frame in the dataset $\mathcal{D}$ to obtain the full set of pseudo-labels $\mathcal{B}^{(0)}$.
This process follows a classical pipeline 
prior to the deep-learning era \cite{douillard2011segmentation, himmelsbach2010fast}. 
We make simple choices for each module:
for ground removal, we fit a linear ground plane similar to \cite{detect-mobile-objects} and remove the estimated ground points before clustering; 
for clustering, we use DBSCAN \cite{dbscan}; 
for bounding box fitting of each point cluster, we use an off-the-shelf algorithm \cite{bbox-fitting-algo}.
While much more sophisticated choices exist, those noisy initial labels are already
good enough to kickstart our learning process.

In the near range (typically within 40m for a LiDAR with 64 beams), objects usually have clear point clusters (see Fig.~\ref{fig:hook}), 
so the clustering labels suffice the bootstrapping purpose.
However, in the long range, each object has much fewer points, making clustering results unreliable, 
so we should rely on neural net's ability to generalize even for initial label generation. 
Consequently, our goal is to \textit{train a detector on near-range only,
such that it can zero-shot generlize to longer range at test time.}
We utilize equivariance of convolutional nets to achieve this goal.
Following~\cite{pointpillars,yang2018pixor}, we train a single-stage CNN detector
on near-range only, where the inputs to the CNN are the voxelized point clouds. 
The voxelizer produces a binary voxel occupancy map $I \in \{0,1\}^{L\times W\times D}$, 
where $L \times W$ defines the input BEV grid resolution, 
and $D$ is the number of channels in the height dimension.
We use a ResNet \cite{resnet} backbone with FPN \cite{fpn} to produce $4$x downsampled feature map, from which a simple convolutional header decodes the box parameters $(x, y, l, w, \theta)$ and the confidence $c$ in a dense fashion (\emph{i.e.}, per BEV pixel).
After this, the set of object detections is obtained by applying confidence thresholding and non-maximum suppression (NMS).
During training, the inputs to the backbone are near-range LiDAR BEV images of resolution $L\times W\times D$; 
during label generation (\emph{i.e.}, at inference), the inputs are longer-range with size $\tilde{L}\times \tilde{W}\times D$, where $\tilde{L} > L$ and $\tilde{W} > W$. 

Considering the difference in point density between near-range and long-range, 
we also propose to randomly drop rays during training, such that the convolutional kernels 
perform better on sparse regions, which is useful for the zero-shot generalization to long-range. This ray-dropping data augmentation is implemented by 
first randomly dropping LiDAR beams, and then randomly dropping points within evenly spaced rows and columns under a range-view image in spherical coordinates.
This process aims to mimic how LiDAR beams tend to be more spaced out as range increases.
We find that ray dropping helps CNNs generalize to longer range significantly better 
when only trained on near range, whose results outperform those from CNNs directly trained on full-range clustering labels.
\begin{align*}
    \text{CNN}^{(0)} &\leftarrow \text{Train}\Big(\Big\{\text{RayDrop}(p_{i})\Big\}_{i \in \mathcal{D}}, \mathcal{B}^{(0)} \mid \text{near-range}\Big) \\ 
    \mathcal{B}^{(1)} &\leftarrow \Big\{ \text{NMS}\Big(\text{CNN}^{(0)}(\text{Voxelize}(p_{i})) \mid \text{full-range} \Big) \Big\}_{i \in \mathcal{D}}
    \label{eq:iter1-boxes}
\end{align*}
The detection outputs above a certain confidence threshold will become our first iteration of self-training pseudo-labels $\mathcal{B}^{(1)}$, which are already able to discover many missed detections and remove false positives from initial point clusters $\mathcal{B}^{(0)}$. This self-correction phenomenon is also observed in~\cite{detect-mobile-objects}, and is a result of the model's limited capacity to overfit to the inconsistent noises in the pseudo-labels.

\subsection{Track, Refine, Retrain, and Repeat: Teaching Unsupervised Detectors to Self-Improve}

Albeit better than the initial labels $\mathcal{B}^{(0)}$, 
the labels $\mathcal{B}^{(1)}$ after the initial training and range extension are far from perfect and require further refinement. 
We propose a framework \textit{Track, Refine, Retrain, and Repeat} to teach unsupervised detectors to self-improve. 
Our framework consists of the following steps: run unsupervised tracking, refine the long tracks, discard the short tracks, re-train the detector on refined pseudo-labels, and repeat the process with thresholded detector outputs as the next round of pseudo-labels. 
The method uses temporal consistency as self-supervision to improve its own pseudo-labels and detection results. We will first describe our unsupervised tracker, and then discuss how we leverage the temporal information stored in tracklet states to refine and iterate.
\vspace{-10pt}
\paragraph{Unsupervised Tracking:}
We follow the tracking-by-detection paradigm, where object detections are first obtained independently for each frame in a temporal sequence, and then discrete association between two consecutive time steps is performed iteratively to form the tracks.
To do so, we employ a simple parametric online tracker. At each time step $t$ in a sequence of frames, each object $j$ has a tracklet $\mathbf{s}_{t}^{j}$ that
stores the estimated state trajectories $\mathbf{S}_{t}=\{\mathbf{s}_{t}^{j} = (\mathbf{b}, v_{x}, v_{y}, c)_{t}^{j}, {n}_{t}^{j}\}$ 
where $\mathbf{b}$ corresponds to the previously introduced box parameters, $(v_x, v_y)$ is the 2D velocity, ${n}_{t}$ is the length of tracklet so far, and $c$ is the confidence score of the tracklet. Given pseudo-labels $\mathcal{B}^{(k)}$ from self-training iteration $k$, we first forecast the new states of each tracklet and then match detections to predictions:
\begin{align*}
    \mathbf{M}_{t} \leftarrow \text{Match}(\mathbf{B}_{t}^{(k)}, \text{Forecast}(\mathbf{S}_{t-1}) )
\end{align*}
We adopt the simplest strategy for forecasting and matching: forecasting assumes constant velocity between frames, and matching is greedy based on bounding box centroid distances. %
For a pseudo-label matched to tracklet $j$ with tracklet length $n_t^j$ so far, we set $m_{t}=n_t^j$, where $m_t$ denotes the \textit{temporal consistency score} of the pseudo-label; if a pseudo-label is unmatched, we set $m_{t}=0$. 
Meanwhile, the online tracker computes its current state $\mathbf{S}_{t}$ based on $\mathbf{S}_{t-1}$ and $\mathbf{M}_{t}$,
which involves adding detections to their matched tracklets, growing unmatched tracklets (to handle occlusions and flickering detections), 
initializing new tracklets for unmatched detections, deleting tracklets that have been unmatched for too long,
and incrementing each tracklet's length $n_{t}^{j}$. 
Finally, we simply run this tracker in both temporal directions,
obtaining bi-directional tracklets $\overrightarrow{\mathbf{S}}_{t}$ and $\overleftarrow{\mathbf{S}}_{t}$. 
Note that each pseudo-label has also stored bi-directional temporal consistency scores $\overrightarrow{m}_{t}^{j}$ and $\overleftarrow{m}_{t}^{j}$, and we aggregate them with $\tilde{m}_{t}^{(k)} = \max(\overrightarrow{m}_{t}^{(k)}, \overleftarrow{m}_{t}^{(k)})$.
\vspace{-10pt}

\paragraph{Pseudo-Label Refinement:}
Based on the temporal consistency scores $\tilde{m}_t^{(k)}$, we divide pseudo-labels into short vs. long tracklets by a threshold $q$. For the short tracklets with $\tilde{m}_{t}^{(k)} < q$, we ignore them in the next round of re-training. For the long tracklets with $\tilde{m}_{t}^{(k)} \geq q$, we refine them in a simple process. Due to the tightest-box-fitting nature of the seed clustering labels, the bbox sizes in $\mathcal{B}^{(k)}$ are smaller for partially observed objects, especially for those farther away from the ego-vehicle. To address this, for each actor trajectory, we set the new length and width of the actor as the top $r$ percentile of all lengths and all widths detected throughout the tracklet. Next, following the corner-based alignment strategy from~\cite{auto4d}, we find the bbox corner that is the closest to the ego-vehicle, with the intuition that this corner is the most observed and likely to be the most reliable. We then update the bbox center with the anchored corner, the original bbox heading and the new bbox size. Fig.~\ref{fig:refinement} illustrates the refinement process.

\paragraph{Iterative Self-Training:}
Following the refinement process, we obtain the refined pseudo-labels $\tilde{\mathbf{B}}_{t}^{(k)}$ and use them to guide the next round of training:
\begin{align*}
    &\tilde{\mathbf{B}}_{t}^{(k)} = \begin{cases}
        \text{Refine}(\overrightarrow{\mathbf{S}})_{t}^{(k)} & \tilde{m}_{t}^{(k)} \geq q \\
        \emptyset & \text{otherwise}
      \end{cases}\\
    &\text{CNN}^{(k)} \leftarrow \text{Train}\Big(\Big\{\text{RayDrop}(p_{i})\Big\}_{i \in \mathcal{D}}, \tilde{\mathcal{B}}^{(k)} \mid \text{full-range} \Big) \\
    &\mathcal{B}^{(k+1)} \leftarrow \Big\{ \text{NMS}\Big(\text{CNN}^{(k)}(\text{Voxelize}(p_{i})) \mid \text{full-range} \Big) \Big\}_{i \in \mathcal{D}}
\end{align*}
where $\emptyset$ indicates that training loss is not applied at the location of this pseudo-object during detector re-training, \emph{i.e.}, we only re-train on pseudo-labels that are temporally consistent and refined. With the re-trained detector, we derive the new generation of pseudo-labels $\mathcal{B}^{(k+1)}$, which can be further refined and used for re-training.

In summary, our \textit{Track, Refine, Retrain, and Repeat} framework constructs an object discovery loop 
where the detector is iteratively re-trained on pseudo-labels of increasingly higher quality as self-training goes on.

%% file: sec/exp.tex
\section{Experimental Evaluation}

\vspace{-5pt}
In this section, we first outline the datasets and metrics we use, including a newly proposed detection metric based on distance-to-collision.
Next, we show that our method outperforms state-of-the-art unsupervised detection methods.
Finally, we present quantitative and qualitative insights into our contributions with a thorough ablation study of different components.

\vspace{-12pt}
\paragraph{Datasets and Experiment Setting:} 
We use Pandaset \cite{xiao2021pandaset} and Argoverse 2 Sensor \cite{wilson2021argoverse} (AV2 for short) to evaluate our method on two different sensor suites.
Pandaset consists of 103 snippets of 8 seconds each, recorded in dense urban traffic in San Francisco as well as the El Camino Real highway. For our experiments, we use the spinning Pandar64 LiDAR.
It features 28 different annotation classes, which we group into 3 main categories: vehicles, cyclists and pedestrians. No annotations are used during training; at test time, we conduct class-agnostic evaluation on three main classes combined, 
since our goal is to train an unsupervised object detector that detects any object.
We split the dataset into 73 training and 30 validation snippets. Both splits evenly distribute across both San Francisco and El Camino Real.
The AV2 dataset is collected in six distinct cities in the U.S. cities with diverse weather (from snowy to sunny). It 
consists of 850 snippets, each 15 seconds long.
The LiDAR data comes from two 32-beam lidars, spinning at 10 Hz in the same direction, but separated in orientation by 180$^\circ$.
We use the official train and validation split with 700 and 150 snippets respectively. 
For our detection models, we focus on the front-range setting with a region of interest (ROI) of $[0, 80]$ meters longitudinally and $[-40, 40]$ meters laterally with respect to the traveling direction of the ego vehicle. The selected ROI allows us to evaluate both near-range and long-range detections.

\vspace{-10pt}
\paragraph{Implementation Details:} For label refinement, we use track length $q = 6$ for both Pandaset and AV2. To update bounding box sizes in long tracklets, we use $r=100\%$ for Pandaset and $r=95\%$ for AV2. For Pandaset we run three rounds of self-training, with long tracklet refinement only applied in the last training round. For AV2 we employ two rounds of self-training, with long tracklet refinement in every training round. More details of the detection model, the tracker, and training are included in the supplementary.

\input{tables_new_dtc/allclass_main_pandaset.tex}
\input{tables_new_dtc/allclass_main_argo.tex}

\begin{figure}
  \centering
  \vspace{-5pt}
  \includegraphics[width=0.95\columnwidth]{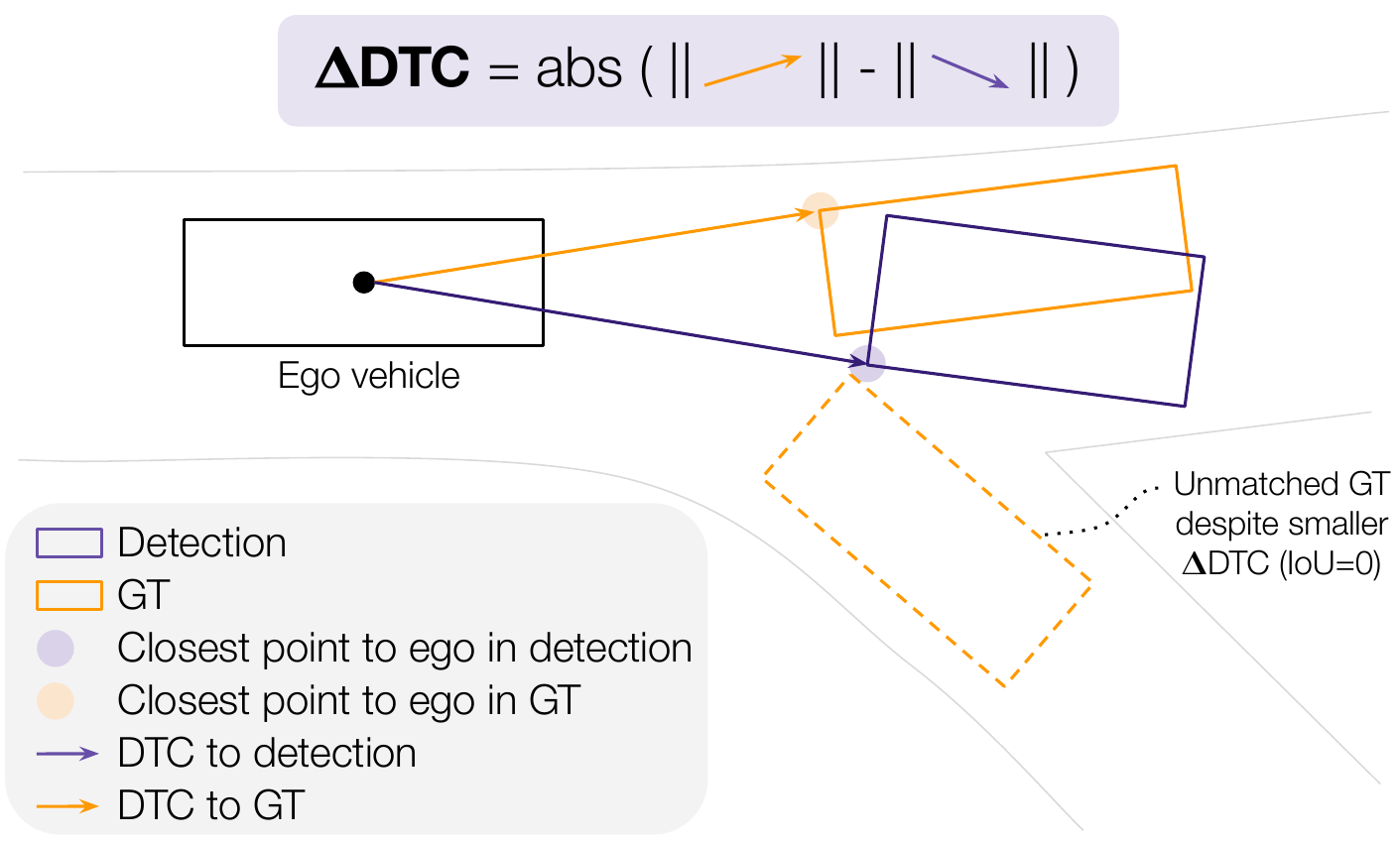}
  \vspace{-10pt}
  \caption{
      \textbf{Our novel matching criteria for detection metrics based on distance-to-collision (DTC)}.
      It prioritizes detecting the closest corner to the SDV correctly, as opposed to the full extent.
  }
  \label{fig:delta-dtc}
  \vspace{-10pt}
\end{figure}

\input{figs/qualitative-comparison.tex}

\vspace{-10pt}
\paragraph{Metrics:}
Our metrics focus on the goal of evaluating a class-agnostic object detector that is capable of detecting any object. 
The commonly used IoU (intersection-over-union) tends to overly penalize incorrect estimation of one side of object dimensions that is intrinsically difficult to predict due to imperfect observation, but we primarily care about accurate predictions of how far the detected objects are from the self-driving vehicle.
After combining annotations from all classes (vehicle, pedestrian, and cyclist), we propose the following evaluation steps to give a full picture about the performance of different methods: 
\begin{itemize}
\item Measure Average Precision (AP) and Recall. AP is a more complete metric, measuring the area under the Precision Recall curve. However, we may identify properly detected objects as false positives due to the absence of annotations for objects in that class. On the other hand, Recall does not penalize false positive detections, making it more suitable for our experimental setting. Since arbitrarily large Recall can be achieved by producing a very large and diverse set of object detections, we evaluate all methods limiting the budget of detections to the 100 most confident ones.
\item Compute these metrics for two different matching criteria, one based on the standard intersection over union (IoU) --- which captures how well we predict the size of the objects, and a novel metric based on the difference in distance-to-collision ($\Delta$DTC) --- which measures how well we can predict how close objects are to the ego vehicle. \cref{fig:delta-dtc} shows our proposed matching criteria based on $\Delta$DTC. For a match between a pair of detection and ground-truth bounding boxes, we require that the two are under a certain $\Delta$DTC threshold (in meters). Subject to this constraint, we perform IoU-based greedy matching to prevent associating detections with ground-truth bounding boxes corresponding to nearby actors (e.g., the dotted bounding box in the adjacent lane), as those may have a corner that is closer to the detection, but have no overlap.
\end{itemize}
Note that we only consider class-agnostic metrics rather than per-class metrics, since our task is to detect any object rather than predict their classes. As a result, naively computing per-class AP will incorrectly penalize true detections for other classes as false positives. Per-class Recall will be artificially low for rarer classes, since constraining the number of detections is often necessary for recall calculation (otherwise 100\% recall is achieved by outputting detections everywhere), and highly represented classes in the data (\emph{e.g.}, vehicles) tend to overwhelm the rest of the classes within a fixed number of detections.

\vspace{-10pt}
\paragraph{Baselines:}
Following~\cite{detect-mobile-objects}, we compare against the following unsupervised baselines. 
\textit{\dbscan{}} \cite{dbscan} performs density-based spatial clustering, grouping points with many nearby neighbors together.
\textit{\dbscantrain{}} learns an object detector with the supervision of DBSCAN labels.
\textit{\dbscanst{}} adds two additional rounds of self-training after the initial training, where the thresholded detection outputs of the object detector are used as pseudo-labels for self-training.
\textit{\ppscore{}} \cite{detect-mobile-objects} estimates the persistence of a point by measuring the variance in the number of LiDAR points within the point's neighborhood encountered in other observations of the same region. 
These PP-scores can then be used as a feature for clustering, retrieving mobile objects that have a low PP-score.
Note that in the original paper multiple traversals over the same area are assumed, which pose a very strict requirement on data collection. To avoid imposing this strict requirement, we only consider a single traversal with multiple observations over a short period of time (\emph{i.e.}, the length of the snippet).
\textit{\ppscoretrain{}} trains a detector on the persistence-based clustering labels with one iteration of training.
Finally, \textit{\modest{}} adds two rounds of self-training after the initial training, where at each round the pseudo-labels coming from the latest self-trained model are filtered using PP scores to discard persistent clusters. 

\input{tables_new_dtc/allclass_ablation.tex}

\input{figs/ssl-label-evolution.tex}

\vspace{-10pt}
\paragraph{Benchmark against state-of-the-art:}
\cref{table:main-all-class-pandaset,table:main-all-class-argo} show a comparison with state-of-the-art unsupervised methods in Pandaset and Argoverse V2 Sensor datasets, respectively. 
The raw \dbscan{} and \ppscore{} labels perform poorly since confidence scores are not output by these methods for ranking during metrics computation. Training with \initlabel{} improves the metrics for both \dbscan{} and \ppscore{}. 
However, additional rounds of self-training lowered the AP for both IoU and DTC metrics.
The results imply that \modest{} relies very heavily on multiple traversals over the same region.
Our method \textit{\ourmethod} outperforms previous methods by a large margin across all IoU and distance thresholds, both for average precision and recall.
\cref{fig:qualitative-comparison} shows that our method can overcome failure modes exhibited by the baselines such as false negative detections for static objects, high density of false negatives at long-range, and some localization and size estimation errors.

\vspace{-10pt}
\paragraph{Effect of the initial training range and ray-dropping:} 
Since point clustering labels tend to be more reliable in the near range due to high point density, we find it beneficial to train on near-range LiDAR images with near-range clustering labels at first, and then rely on the translation-invariance property of ConvNets to generalize zero-shot to longer range. 
This is empirically demonstrated by the improvement from $M_1 \to M_2$ in \cref{table:pandaset-ablations}, where the zero-shot generalization from short-range (0-40m) to full-range (0-80m) ($M_2$) performs much better than directly training on full-range ($M_1$).
We also find that random ray dropping as a data augmentation technique during the initial training ($M_2 \to M_3$) can help the ConvNet generalize better to longer range in terms of detecting more objects, although it seems to slightly sacrifice localization accuracy as shown by the metrics evaluated at high IoU values, likely because ray dropping encourages the detector to focus more on one side of objects with the most points facing the ego vehicle rather than the overall object shapes. 

\vspace{-10pt}
\paragraph{Effect of short tracklet filtering:}
Using unsupervised tracking to ignore temporally inconsistent \initlabel{} is important.
This makes the pseudo-labels less noisy, providing better supervision to the model.
Training with these filtered set of labels is beneficial, as shown by the improvement from $M_3 \to M_4$ in \cref{table:pandaset-ablations}.

\vspace{-10pt}
\paragraph{Effect of self-training loop and long tracklet refinement:}
\cref{fig:ssl-label-evolution} visually shows the evolution of the detections through self-training (loop shown in \cref{fig:overall-pipeline}). Initially, point clustering labels are quite noisy, identifying many background regions as objects.
After the first training iteration, the model can clearly distinguish foreground from background (\emph{i.e.}, removes false positives), and discovers other vehicles missed by clustering.
Further rounds of self-training discover additional objects such as vehicles and cyclists, and increases the object localization accuracy.
Quantitatively, the effect of 3 rounds of self-training is shown in \cref{table:pandaset-ablations}, where we can see $M_5$ and $M_6$ clearly outperform the other ablations.
The difference between $M_5$ and $M_6$ is that in $M_5$ we skip the long tracklet CornerAlign refinement stage, going directly from tracking and short tracklet filtering to re-training. In $M_6$ we refine the long tracklets in the pseudo-labels provided by $M_5$, and add one more round of self-training with the refined pseudo-labels. We can see that the refinement stage is able to further improve the IoU metrics. Note that the DTC metrics dropped with long tracklet refinement, most likely because increasing the bounding box size with inaccurate heading negatively impacts the accuracy of the closest distance to the ego-vehicle. This showcases the importance of having both IoU and DTC metrics, as they focus on different aspects of detection accuracy.

%% file: tables_new_dtc/allclass_main_pandaset.tex
\begin{table*}
    \centering
    \vspace{-10pt}
    \begin{tabularx}{\textwidth}{l   s s s   s s s   s s s   s s s}
      \toprule
      {} & \multicolumn{3}{c}{AP @ IoU} & \multicolumn{3}{c}{Recall  @ IoU} & \multicolumn{3}{c}{AP @ $\Delta$DTC (m)} & \multicolumn{3}{c}{Recall @ $\Delta$DTC (m)} \\
      \cmidrule(lr){2-4} \cmidrule(lr){5-7} \cmidrule(lr){8-10} \cmidrule(lr){11-13}
      {} & 0.3 & 0.5 & 0.7 & 0.3 & 0.5 & 0.7      & 1.5 & 1.0 & 0.5 & 1.5 & 1.0 & 0.5  \\
      \midrule
      DBSCAN \cite{dbscan}  & 3.5 & 1.1 & 0.3 & 28.6 & 15.9 & 8.3 & 10.9 & 9.9 & 8.0 & 50.9 & 48.3 & 43.1  \\
      \dbscantrain{}      & 21.6 & 12.0 & 6.4 & 42.0 & 26.0 & 13.8 & 41.3 & 39.3 & 34.2 & 71.8 & 69.4 & 62.1  \\
      \dbscanst{} \cite{detect-mobile-objects} & 20.1 & 12.3 & 6.3 & 42.2 & 26.1 & 13.5 & 39.2 & 37.4 & 30.1 & 71.8 & 69.4 & 61.4 \\
      PP score \cite{detect-mobile-objects}  & 6.4 & 2.0 & 0.4 & 32.8 & 18.7 & 8.8 & 14.0 & 12.2 & 9.2 & 48.5 & 45.0 & 38.9 \\
      \ppscoretrain{} \cite{detect-mobile-objects}   & 28.4 & 14.0 & 4.9 & 47.4 & 26.6 & 12.5 & 42.2 & 38.9 & 31.7 & 69.5 & 66.0 & 57.5 \\
      MODEST \cite{detect-mobile-objects} (1 traversal)  & 22.8 & 7.5 & 2.8 & 49.7 & 28.9 & 14.9 & 38.8 & 36.4 & 30.2 & 70.2 & 66.9 & 59.0 \\
      Ours           & \textbf{43.5} & \textbf{29.5} & \textbf{18.1} & \textbf{62.8} & \textbf{44.8} & \textbf{28.1} & \textbf{51.8} & \textbf{48.8} & \textbf{41.1} & \textbf{75.7} & \textbf{72.3} & \textbf{63.7} \\
      \bottomrule
    \end{tabularx}
    \vspace{-5pt}
    \caption{
      \textbf{[Pandaset] Comparison against state-of-the-art}. Evaluated on the range 0-80m on all classes (vehicle, pedestrian, cyclist). %
    }
    \label{table:main-all-class-pandaset}
  \end{table*}

%% file: tables_new_dtc/allclass_main_argo.tex
\begin{table*}
  \centering
  \begin{tabularx}{\textwidth}{l   s s s   s s s   s s s   s s s}
    \toprule
    {} & \multicolumn{3}{c}{AP @ IoU} & \multicolumn{3}{c}{Recall  @ IoU} & \multicolumn{3}{c}{AP @ $\Delta$DTC (m)} & \multicolumn{3}{c}{Recall @ $\Delta$DTC (m)} \\
    \cmidrule(lr){2-4} \cmidrule(lr){5-7} \cmidrule(lr){8-10} \cmidrule(lr){11-13}
    {} & 0.3 & 0.5 & 0.7 & 0.3 & 0.5 & 0.7      & 1.5  & 1.0 & 0.5 & 1.5 & 1.0 & 0.5  \\
    \midrule
    DBSCAN \cite{dbscan}                                  & 2.1 & 1.0 & 0.4 & 26.4 & 17.9 & 10.9 & 6.8 & 6.4 & 5.4 & 47.2 & 45.7 & 41.9 \\
    \dbscantrain{}                                     & 15.5 & 10.7 & 5.7 & 29.7 & 19.3 & 10.1 & 27.6 & 26.6 & 23.3 & 58.0 & 56.6 & 51.7 \\
    \dbscanst{} \cite{detect-mobile-objects}  & 12.5 & 9.1 & 5.5 & 30.1 & 19.6 & 10.3 & 23.9 & 23.1 & 20.4 & 58.4 & 57.0 & 52.2 \\
    PP score \cite{detect-mobile-objects}                 & 5.5 & 2.5 & 0.9 & 33.7 & 22.8 & 13.8 & 11.6 & 10.4 & 8.4 & 49.1 & 46.5 & 41.6 \\
    \ppscoretrain{} \cite{detect-mobile-objects}      & 24.4 & 15.8 & 7.4 & 43.9 & 27.4 & 14.2 & 38.9 & 36.8 & 31.2 & 71.5 & 68.0 & 59.5 \\
    MODEST \cite{detect-mobile-objects} (1 traversal)     & 20.6 & 9.9 & 2.7 & 47.0 & 28.6 & 13.0 & 38.6 & 35.9 & 28.1 & 74.0 & 70.4 & 61.0 \\
    Ours                                                  & \textbf{35.4} & \textbf{24.5} & \textbf{12.9} & \textbf{55.1} & \textbf{37.5} & \textbf{21.4} & \textbf{46.4} & \textbf{43.9} & \textbf{38.1} & \textbf{74.5} & \textbf{71.0} & \textbf{63.1} \\  %
    \bottomrule
  \end{tabularx}
  \vspace{-5pt}
  \caption{
    \textbf{[AV2] Comparison against state-of-the-art}. Evaluated on the range 0-80m on all classes (vehicle, pedestrian, cyclist). \dbscanst{}, MODEST and ours all employ two additional rounds of self-training after the initial training.
  }
  \label{table:main-all-class-argo}
  \vspace{-10pt}
\end{table*}

%% file: figs/qualitative-comparison.tex
\begin{figure*}
    \vspace{-15pt}
    \centering
    \includegraphics[width=0.95\linewidth]{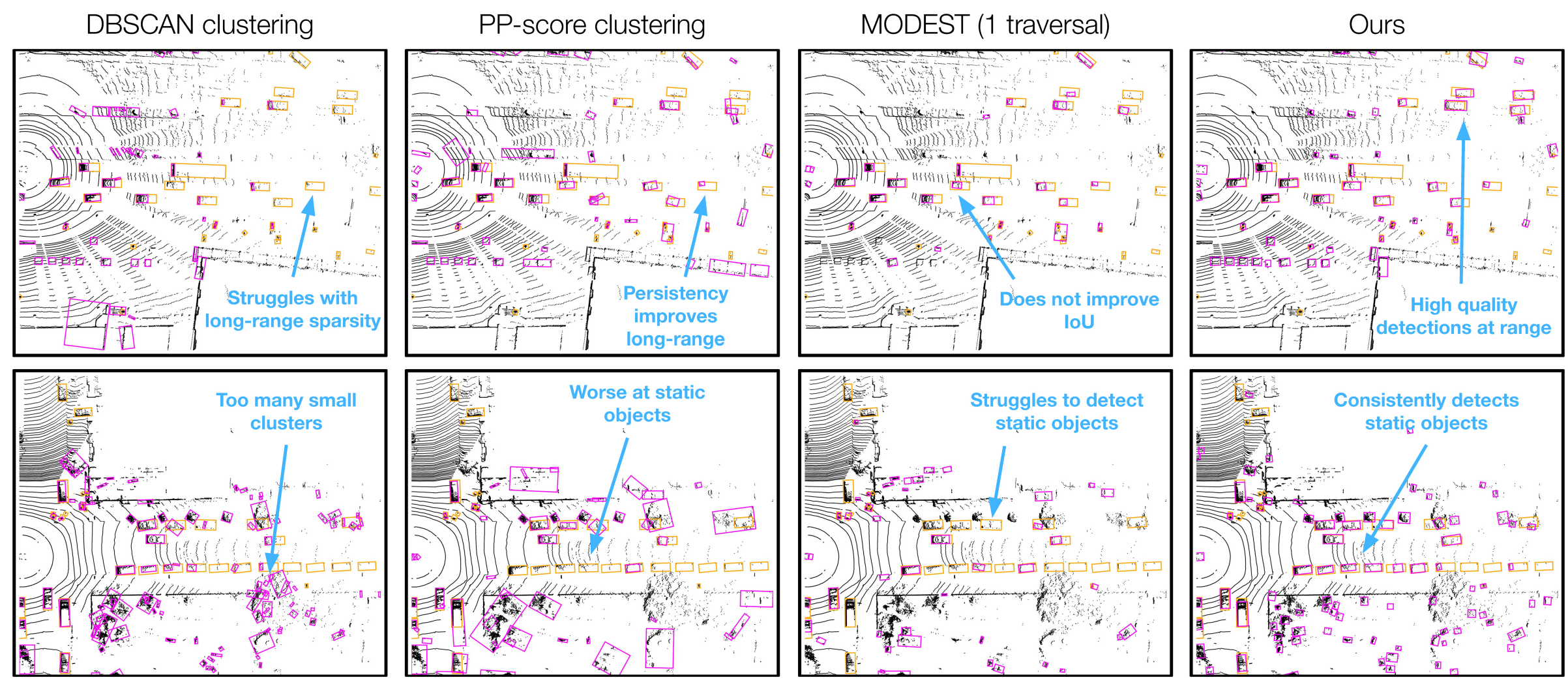}
    \vspace{-10pt}
    \caption{
        \textbf{[Pandaset] Qualitative results:} clustering algorithms (DBSCAN and PP-score), state-of-the-art (MODEST) and our method. We show detections in purple and ground-truth bboxes in orange.
    }
    \label{fig:qualitative-comparison}
    \vspace{-15pt}
  \end{figure*}

%% file: tables_new_dtc/allclass_ablation.tex
\begin{table*}
    \centering
    \vspace{-10pt}
    \begin{tabularx}{\textwidth}{s s s s s s  s s s   s s s   s s s   s s s}
      \toprule
      {} & {} & {} & {} & {} & {} & \multicolumn{3}{c}{AP @ IoU} & \multicolumn{3}{c}{Recall  @ IoU} & \multicolumn{3}{c}{AP @ $\Delta$DTC} & \multicolumn{3}{c}{Recall @ $\Delta$DTC} \\
      \cmidrule(lr){7-9} \cmidrule(lr){10-12} \cmidrule(lr){13-15} \cmidrule(lr){16-18}
      ID & \textcolor{BurntOrange}{ITR} & \textcolor{ForestGreen}{RD} & \textcolor{TealBlue}{SRef} & \textcolor{Fuchsia}{ST2} & \textcolor{Rhodamine}{LRef} & 0.3 & 0.5 & 0.7 & 0.3 & 0.5 & 0.7      & 1.5 & 1.0 & 0.5 & 1.5 & 1.0 & 0.5 \\
      \midrule
      $M_1$ & 80 &            &            &            &            & 21.6 & 12.0 & 6.4 & 42.0 & 26.0 & 13.8 & 41.3 & 39.3 & 34.2 & 71.8 & 69.4 & 62.1 \\
      $M_2$ & 40 &            &            &            &            & 23.2 & 13.0 & 5.3 & 42.9 & 25.7 & 12.8 & 41.8 & 39.5 & 33.9 & 72.0 & 69.5 & 62.1 \\
      $M_3$ & 40 & \checkmark &            &            &            & 26.6 & 14.5 & 4.1 & 44.6 & 24.5 & 10.0 & 46.1 & 43.4 & 36.5 & 72.0 & 69.0 & 60.8 \\
      $M_4$ & 40 & \checkmark & \checkmark &            &            & 32.3 & 17.8 & 7.5 & 51.6 & 31.9 & 15.4 & 48.7 & 46.1 & 40.3 & 74.1 & 71.4 & 64.6 \\
      $M_5$ & 40 & \checkmark & \checkmark & \checkmark &            & \textbf{43.5} & 26.2 & 13.2 & 58.6 & 38.4 & 20.6 & \textbf{55.2} & \textbf{52.3} & \textbf{45.2} & \textbf{77.1} & \textbf{73.9} & \textbf{65.7}  \\
      $M_6$ & 40 & \checkmark & \checkmark & \checkmark & \checkmark & \textbf{43.5} & \textbf{29.5} & \textbf{18.1} & \textbf{62.8} & \textbf{44.8} & \textbf{28.1} & 51.8 & 48.8 & 41.1 & 75.7 & 72.3 & 63.7  \\
      \bottomrule
    \end{tabularx}
    \vspace{-10pt}
    \caption{
        \textbf{[Pandaset] Ablation study}. All class evaluation in the range 0-80m.
        Legend: ID=Model identifier, \textcolor{BurntOrange}{ITR=Initial training range (first iteration) from 0 to X meters}, \textcolor{ForestGreen}{RD=Ray-dropping}, \textcolor{TealBlue}{SRef=Tracking \& short tracklet filtering}, \textcolor{Fuchsia}{ST2=2 rounds of additional self-training}, \textcolor{Rhodamine}{LRef=Long tracklet refinement}. 
    }
    \label{table:pandaset-ablations}
  \end{table*}

%% file: figs/ssl-label-evolution.tex
\begin{figure*}
    \centering
    \vspace{-10pt}
    \includegraphics[width=0.95\linewidth]{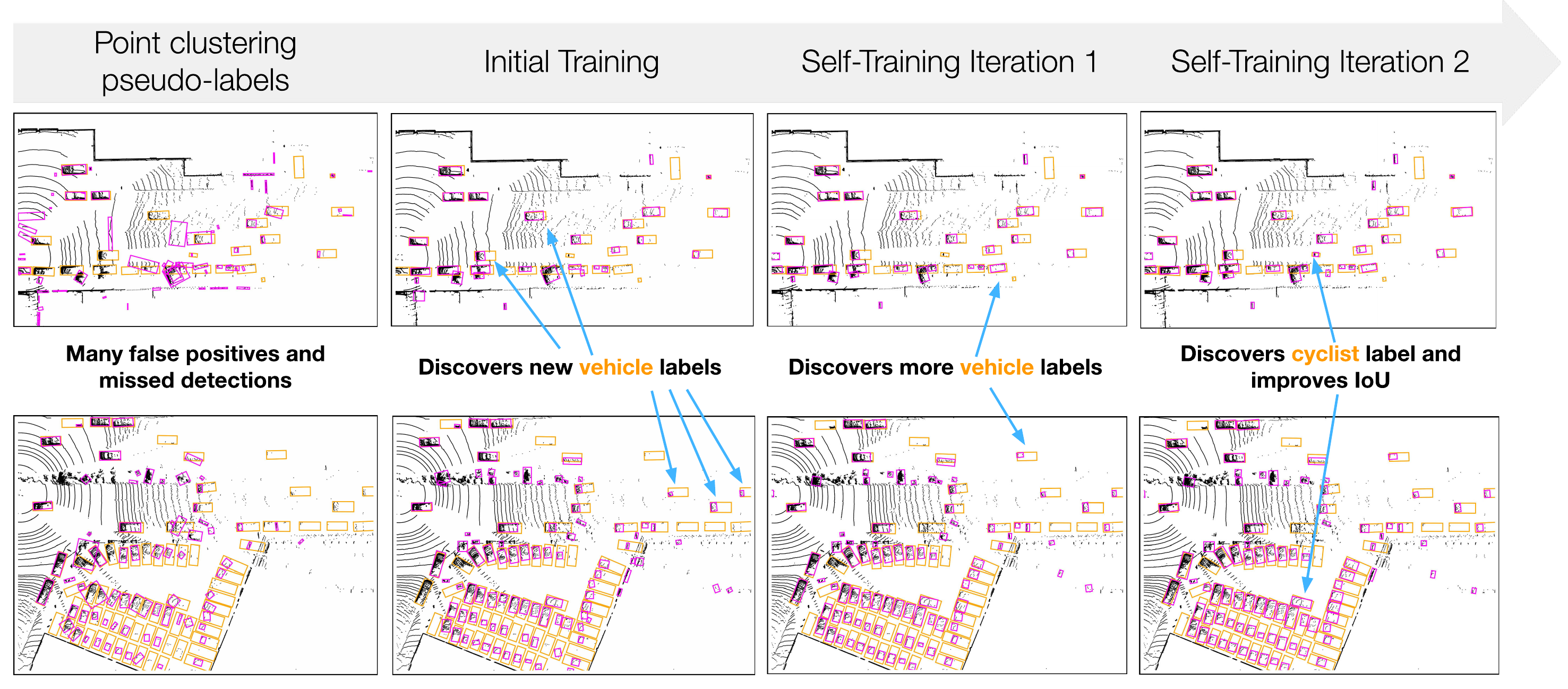}
    \vspace{-10pt}
    \caption{\textbf{[Pandaset] Evolution of self-training labels}: with initial training and more rounds of self-training, we are able to remove false positives, discover missed detections, and improve bbox accuracy. We show detections in purple and ground-truth bboxes in orange.}
    \label{fig:ssl-label-evolution}
    \vspace{-10pt}
  \end{figure*}

%% file: sec/conclusion.tex
\section{Conclusion}

We have proposed a novel method, \textit{\ourmethod}, for unsupervised object detection from LiDAR point clouds. Using weak object priors (near-range point clustering) as a bootstrapping step, our method can train an object detector with no human annotations, by first utilizing the translation equivariance of CNNs to generate long-range pseudo-labels, and then deriving self-supervision signals from the temporal consistency of object tracks. Our proposed self-training loop is highly effective for teaching an unsupervised detector to self-improve. We validate our results on two real-world datasets, Pandaset and Argoverse 2 Sensor, where our model outperforms prior unsupervised methods by a significant margin. Making self-supervised learning work on real-world robot perception is an exciting challenge for AI, and our work takes a step towards allowing robots to make sense of the visual world without human supervision. 

%% file: main_supp.tex
%% LaTeX2e file `main_supp.tex'
%% generated by the `filecontents' environment
%% from source `PaperForReview' on 2023/06/16.
%%

In the appendix, we first describe additional implementation details in Sec.~\ref{sec:supp-impl-details}. We next showcase additional quantitative results in Sec.~\ref{sec:supp-quant} and additional qualitative results in Sec.~\ref{sec:supp-qual}. For videos of our qualitative results, visit the project website: \url{https://waabi.ai/research/oyster}.
\input{sec/supp/impl_details}

\input{sec/supp/quantitative_results}

\input{sec/supp/qualitative_results}

%% file: sec/supp/impl_details.tex
\vspace{-1pt}
\section{Implementation Details}
\label{sec:supp-impl-details}

\vspace{-1pt}
\paragraph{Detector Architecture:}

We use a detector similar to single-stage PIXOR detector \cite{yang2018pixor}. The inputs to the detector are 3D voxelized binary LiDAR images from Bird-Eye View (BEV), with front-range region of interest (ROI) of $[0, 80]$ meters longitudinally and $[-40, 40]$ meters laterally with respect to the traveling direction of the ego vehicle. The step size of the voxel in BEV is $0.15625$ meters, producing an input feature resolution of $512\times 512$. In the $z$-axis of the voxels, we clip the range to be $[-1.5, 5.5]$ meters, and use a step size of $0.2$ meters. We also concatenate $4$ previous voxelized frames to the current frame along the channel dimension, after ego motion compensation has been applied to the LiDAR points of those $4$ previous frames (assuming access to the poses of ego vehicle in the world coordinate per frame). As a result, the channel dimension for the input feature maps is $35\times 5=175$. Unlike PIXOR, we do not use the point-wise reflectance values. 

The detector architecture is similar to ResNet \cite{resnet} with Feature Pyramid Network (FPN) \cite{fpn}. The ResNet backbone has: initial stem layers which downsample the feature resolution by $2\times$, and then $3$ residual stages, each of which downsamples the feature resolution by another $2\times$. The downsampling in the stem layers is done via strided $\texttt{conv3x3}$ in the first layer of stem; the downsampling in $3$ residual stages is done only inside the first residual block of each stage, via strided $\texttt{conv1x1}$ on the residual inputs and strided $\texttt{conv3x3}$ in the middle of a $\texttt{Bottleneck}$ block. The $3$ residual stages have $(6,6,4)$ blocks respectively. We use Sync BatchNorm ($\texttt{SyncBN}$) \cite{batchnorm} for all normalization layers, and the weights and biases of the final \texttt{SyncBN} inside each \texttt{Bottleneck} block are initialized to be zero for better initialization. We use ReLU activations as all non-linearities. The stem layers reduce the channel dimension of feature volumes from $175$ in the BEV LiDAR inputs to $64$, and the $\texttt{Bottleneck}$ layers in $3$ residual stages subsequently use $(48, 64, 96)$ as the bottleneck dimensions and expand those dimensions by $4\times$ as the outputs of the ResNet backbone and the inputs to the FPN neck. Therefore, FPN inputs have ($4\times$, $8\times$, $16\times$) downsampled resolutions with channels $(192, 256, 384)$, which are combined to produce a feature map with $4\times$ downsampled resolution and $128$ channels. 

Detection is performed on the $4\times$ downsampled resolution from BEV. After FPN, we use separate headers for the classification branch and the regression branch. The classification branch applies $4$ layers of $(\texttt{conv3x3}\rightarrow\texttt{SyncBN}\rightarrow\texttt{ReLU})$ with channel size $48$ and then a \texttt{conv1x1} layer at the end; the regression branch is similar but with channel size $128$. The classification branch produces binary classification logits to predict the presense of class-agnostic objects. The regression branch predicts a 6-dimensional vector $(dx, dy, \log l, \log w, \sin\theta, \cos\theta)$, where $(dx, dy)$ represent the predicted offsets of object centers from the BEV $\texttt{meshgrid}$, and $(\log l, \log w)$ represent the log of the predicted physical lengths of the objects in terms of meters. The outputted values for $(\sin\theta, \cos\theta)$ are unconstrained since we only use $\theta=\texttt{atan2}(\sin\theta, \cos\theta)$ in the prediction and loss calculations. The classification branch uses focal initiailzation for the binary logits \cite{focal-loss}. 

\vspace{-10pt}

\paragraph{Ray-Dropping based Data Augmentations:}

There are two ray-dropping techniques we use as data augmentations to artifically make dense LiDAR point clouds sparser. 

Given a $64$-beam LiDAR and their (ordered) beam IDs, we first randomly drop points based on their beam IDs. More specifically, we randomly sample a beam drop ratio from $[1,2,3]$, then randomly sample a starting beam index between $0$ and this beam drop ratio, and finally, only keep the points whose beam IDs minus the starting beam index is a multiple of the beam drop ratio. In other words, we only keep the points inside beams that are uniformly spaced according to beam IDs, and the spacing is determined by the sampled beam drop ratio. Note that, when beam drop ratio is $1$, the point clouds stay intact. 

Then, we convert the 3D points from Euclidean coordinates to spherical coordinates, discretize the spherical coordinates according to a certain resolution, and randomly subsample the points with uniform spacing in the discretized spherical coordinates. More specifically, we convert the 3D points whose Euclidean norms are bigger than $0.1$ into spherical coordinates $(\texttt{theta}, \texttt{phi}, \texttt{radial\_dist})$, and then randomly sample the discretization resolutions of \texttt{theta} and \texttt{phi} from $[600, 900, 1200, 1500]$. We then randomly sample spherical drop ratios from $[1,2]$ (when this ratio is 1, the point clouds stay intact). We only keep the points inside evenly-spaced discretized spherical coordinates for both rows and columns, and the spacing is determined by the sampled spherical drop ratio. 

We apply the two data augmentations in the sequential order above. Ray dropping is applied not only during initial bootstrapping to help the near-range detector generalize to longer range, but during later rounds of self-training as well when training is directly done on full range. 

\vspace{-10pt}

\paragraph{Point Clustering and Bounding Box Fitting:}

For point clustering, we first remove the points outside the detection ROI. Then, we try to remove the ground points based on a fitted ground plane. To obtain a robust estimate of the ground plane, we perform the first round of DBSCAN \cite{dbscan} point clustering to get the $50$-th percentile $z$-axis (height) of non-outlier points, and then only fit a linear plane on the points below this height. (On Argoverse 2, we change the $50$-th percentile to $30$-th percentile.) For DBSCAN, we use $\texttt{eps}=0.4$ and $\texttt{min\_samples}=8$. The plane fitting module uses Random sample consensus (RANSAC) algorithm \cite{RANSAC} with a minimum sample size of $1000$ to fit a linear plane. To obtain a conservative estimate of the ground, we apply two rounds of ground fitting, with a small negative offset applied to the $z$-axis proportionally and all the points above the fitted plane removed after the first round and before the second round of fitting, similar to \cite{detect-mobile-objects}. We then remove the points below this fitted plane, and then do DBSCAN clustering again with the same parameters. For each cluster, we fit a tightest bounding box using the algorithm described in \cite{bbox-fitting-algo}. For persistence-based clustering \cite{detect-mobile-objects}, we first count the point-wise number of neighbors with maximum threshold $0.5m$ among the adjacent $5$ past frames (including the current frame), and then construct a KNN-based neighborhood graph with maximum number of neighbors equal to $70$ and the distance metric being L1 distance between ephemerality scores \cite{detect-mobile-objects}. DBSCAN for persistence-based clustering uses parameters $\texttt{eps}=0.1$ and $\texttt{min\_samples}=10$. We remove the boxes that either have a BEV area of less than $0.4 m^{2}$, or have a length $l$ or width $w$ bigger than $15m$. We use \textit{the same} ground fitting, point clustering, and boudning box fitting algorithms for our method and all our baselines. 

The near-range training in the initial bootstrapping phase uses an RoI of $[0, 40]$ meters longitudinally and $[-20, 20]$ meters laterally w.r.t the traveling direction of the ego vehicle, since the point clustsers and their fitted bounding boxes tend to have higher quality in this near-range RoI. 

\vspace{-10pt}

\paragraph{Unsupervised Tracking:}
We employ a simple online tracker. For each new frame at time step $t$ with detections $\mathbf{B}_t = \{\mathbf{b}_t^{l}\}$ where each $\mathbf{b}_t^{l} = (x_t^l, y_t^l, l_t^l, w_t^l, \theta_t^l) \in \mathbb{R}^5$ is the individual 2D BEV bounding box, we compute a cost matrix with existing tracklets $\mathbf{S}_t = \{\mathbf{s}_t^j\}$ as follows. For each tracklet $j$, we first predict its bbox position $(x^j_t, y^j_t)$ at time $t$: if the tracklet has at least two past frames, we set $(x^j_t, y^j_t) = 2 * (x^j_{t-1}, y^j_{t-1}) - (x^j_{t-2}, y^j_{t-2})$ via naive extrapolation (assuming constant velocity between two adjacent frames); otherwise we simply set $(x^j_t, y^j_t) = (x^j_{t-1}, y^j_{t-1})$. Then, for each pair of the detected bbox $\mathbf{b}_t^l$ and the predicted tracklet bbox $\mathbf{b}_t^j$, we compute the Euclidean distance between the bbox centroids as $\ell^{j,l} = \sqrt{(x_t^j - x_t^l)^2 + (y_t^j - y_t^l)^2}$. For each existing tracklet, we simply employ a greedy strategy to find the nearest detection $l^\ast = \argmin_{l}{\ell^{j,l}}$, and if the closest distance $l^{j, l^\ast}$ is greater than a threshold of 5.0m, then the tracklet has no match. We use greedy matching instead of a more sophisticated matching strategy such as Hungarian matching because it is more robust to noisy and spurious detections.

If a tracklet is matched to a new detection, we add the detection to the tracklet and update the tracklet score $c_t^j = \frac{w \cdot c_{t-1}^j + c_t^l}{w + 1.0}$, where $c_{t-1}^j$ is the old tracklet score, $c_t^l = 1.0$ is the detection confidence score we set for every new detection, and $w = \sum_{i=1}^{n_{t-1}^j}{0.9^i}$ where $n_{t-1}^j$ is the number of tracking steps in the tracklet.

If a tracklet is not matched, we grow the tracklet by naively extrapolating the position and angle, and set the new confidence score as $c_t^j = 0.9c_{t-1}^j$.

If a new detection is not matched to any tracklet, we start a new tracklet with confidence score $c_t^j = 0.9$. We set the starting confidence score as 0.9 instead of 1.0 so that the longer tracklet will survive the subsequent NMS due to its higher confidence score. 

We terminate all tracklets with a tracking confidence score less than 0.1, and apply NMS at the end over all existing tracklets in the current frame with an IoU threshold of 0.1. We repeat this process for the next frame at time $t + 1$ until the end of the sequence.

As mentioned in the main paper, unsupervised tracking is done in both the forward and the reverse direction of time, to compute the temporal consistency score of a given pseudo-label; pseudo-labels with low consistency scores are ignored. This procedure is applied in every round of self-training and the initial bootstrapping phase as well.

\vspace{-10pt}

\paragraph{Training Details:}

The training loss for the detector is a combination of the focal loss \cite{focal-loss} and the rotation-robust generalized IoU (rgiou) loss \cite{rgiou-loss}. The focal loss uses an \texttt{alpha} of $0.5$ and a \texttt{gamma} of $2.0$; the losses for positive labels and negative labels are computed and summed separately, both normalized by the number of positive labels. To determine which pixels on the feature map (note that the feature map has a $4\times$ downsampled resolution) count as positive or negative, we first compute the axis-aligned intersection-over-union (IoU) for all pixels in the output feature map with respect to every object (in other words, the IoU is calculated between a point and a box with different centers but aligned sizes and yaws). For every object:
\begin{itemize}
	\item If there exist pixels with this IoU value bigger than $0.5$, then we randomly sample one such pixel (per object) to apply the losses on (positive focal loss + rgiou loss), and do not apply any losses on other pixels where this IoU value is also bigger than $0.5$; 
	\item Otherwise, if there are no pixels with this IoU value bigger than $0.5$, then we select the pixel with the highest IoU value (if the highest value is zero, then this label is ignored) to apply the losses on, and do not apply any losses on other pixels where this IoU value is smaller than the selected pixel but bigger than $0.3$. 
	\item For all the pixels that are selected as \textit{positive}, apply positive focal loss + rgiou loss; for all the pixels that are not selected as \textit{positive} and not set to be \textit{ignored}, apply negative focal loss. 
\end{itemize}

We use Adam optimizer \cite{adam-optimizer} with decoupled weight decay \cite{adamw}, with a learning rate of $0.004$ and a weight decay coefficient of $0.0001$. For PandaSet, we train the detector for $20$ epochs without learning rate decay. For Argoverse 2, we train on a $10\times$ subsampled dataset (meaning that for each sequence in the training set, we subsample the number of frames by $10\times$ per sequence to speed up training), and train for 40 epochs, with the learning rate decreased by $10\times$ at the $20$-th epoch. On both datasets, we use a batch size of $8$ per GPU, and apply distributed training with $4$ GPUs. 

We use non-maximum suppression (NMS) in detection. NMS is applied to the top $1000$ scored detections per frame with a threshold of $0.1$. After NMS, we take the top $100$ detections. For pseudo-label generation of the next self-training round, we use a score threshold of $0.4$ after NMS.

%% file: sec/supp/quantitative_results.tex
\section{Additional Quantitative Results}
\label{sec:supp-quant}

\input{tables_new_dtc/allclass_ablation_buckertizer_new.tex}
\input{tables_new_dtc/allclass_supervised_pandaset.tex} 
\input{figs/quantitative_ssl_iter.tex}

\paragraph{Detailed breakdown of near-range and far-range metrics:} To showcase the effect of near-range training and zero-shot extension to far-range and training with ray drop, we additionally break down the [0, 80]m metrics in the first three rows of Tab.~3 in the main paper into near-range ($[0, 40)$m) and far-range ($[40, 80]$m). As shown in Tab.~\ref{table:pandaset-ablations-range}, by training in the near range with ray dropping, the detection metrics for far-range objects significantly improve.
\vspace{-10pt}
\paragraph{Measuring the gap between unsupervised and supervised methods:}
\cref{table:supervised-all-class-pandaset} shows a comparison between our method and multiple supervised counterparts, trained with varying levels of supervision. We evaluate with all classes (vehicle, pedestrian, cyclist) combined, in a class-agnostic manner. The table shows that despite the advancement of unsupervised object detection made in this work, the remaining gap between unsupervised learning and supervised learning for object detection remains large in the context of self-driving. More specifically, our method \textit{\ourmethod} surpasses supervised learning on a single log and starts to match its supervised counterpart on $2$ logs; however, compared to the performance of supervised learning on frames sub-sampled from the whole dataset (in other words, from all the logs), our method is not yet as competitive.
\vspace{-10pt}
\paragraph{Performance vs. number of self-training iterations:} We additionally study how the number of self-training iterations affects unsupervised detection performance. Fig.~\ref{fig:quantitative-ssl-iter} showcases IoU and DTC metrics with respect to the number of self-training iterations on both Pandaset and AV2. Note that in the main paper we apply long tracklet refinement only in the last iteration of the Pandaset experiments. For a more fair comparison, in this ablation study we do not apply the long tracklet refinement in the last iteration so that each Pandaset self-training iteration employs the same pseudo-label post-processing process. On the other hand, similar to the main paper, we apply full refinement (both long tracklet refinement and short tracklet filtering) in every AV2 self-training iteration. Our results show that overall, for both datasets, more self-training iterations improved almost all metrics, with the exception that DTC-based recall dropped slightly with the third self-training iteration. Therefore, we stop the self-training loop at iteration 3 for Pandaset and iteration 2 for AV2. We leave DTC recall improvement with more self-training iterations as future work.

%% file: tables_new_dtc/allclass_ablation_buckertizer_new.tex
\begin{table*}
    \centering
    \begin{tabularx}{\textwidth}{s | s s s s s s s s s s s s s s s}
      \toprule
      \multicolumn{1}{c}{} & {} & {} & {} & \multicolumn{3}{c}{AP @ IoU} & \multicolumn{3}{c}{Recall  @ IoU} & \multicolumn{3}{c}{AP @ $\Delta$DTC} & \multicolumn{3}{c}{Recall @ $\Delta$DTC} \\
      \cmidrule(lr){5-7} \cmidrule(lr){8-10} \cmidrule(lr){11-13} \cmidrule(lr){14-16} 
      \multicolumn{1}{c}{} & ID & \textcolor{BurntOrange}{ITR} & \textcolor{ForestGreen}{RD} & 0.3 & 0.5 & 0.7 & 0.3 & 0.5 & 0.7      & 1.5 & 1.0 & 0.5 & 1.5 & 1.0 & 0.5 \\
      \midrule
      \parbox[t]{2mm}{\multirow{3}{*}{\rotatebox[origin=c]{90}{$[0, 40)$m}}} & $M_1$ & 80 &            & 39.7 & 23.9 & \textbf{12.7} & 62.7 & \textbf{43.6} & \textbf{25.1} & 52.8 & 50.6 & 45.4 & 79.5 & 77.1 & 70.7\\
      {} & $M_2$ & 40 &            & 42.7 & 25.6 & 10.6 & \textbf{64.5} & 43.1 & 22.8 & 55.1 & 52.7 & 46.7 & \textbf{80.4} & \textbf{78.0} & \textbf{71.5}\\
      {} & $M_3$ & 40 & \checkmark & \textbf{44.0} &\textbf{27.0} & 7.6 & 61.7 & 39.1 & 17.0 & \textbf{56.7} & \textbf{54.1} & \textbf{47.0} & 77.5 & 74.8 & 67.4\\
      \midrule
      \parbox[t]{2mm}{\multirow{3}{*}{\rotatebox[origin=c]{90}{$[40, 80]$m}}} & $M_1$ & 80 &            & 2.1 & 0.3 & 0.0 & 19.5 & 7.4 & 2.0 & 25.0 & 23.5 & 18.7 & 61.6 & 59.1 & 51.1\\
      {} & $M_2$ & 40 &            & 2.5 & 0.5 & 0.1 & 19.5 & 7.2 & 2.2 & 22.7 & 20.7 & 16.1 & 61.0 & 58.2 & 50.2\\
      {} & $M_3$ & 40 & \checkmark & \textbf{6.7} & \textbf{1.3} & \textbf{0.3} & \textbf{25.8} & \textbf{9.1} & \textbf{2.6} & \textbf{30.7} & \textbf{28.2} & \textbf{22.0} & \textbf{64.3} & \textbf{61.1} & \textbf{52.0}\\[0.5ex]
      \bottomrule
    \end{tabularx}
    \caption{
        \textbf{[Pandaset] Ablation study with ranges}. All class evaluation in the range 0-40m (top) and 40-80m (bottom).
        Legend: ID=Model identifier, \textcolor{BurntOrange}{ITR=Initial training range (first iteration) from 0 to X meters}, \textcolor{ForestGreen}{RD=Ray-dropping}. 
    }
    \label{table:pandaset-ablations-range}
  \end{table*}

%% file: tables_new_dtc/allclass_supervised_pandaset.tex
\begin{table*}
  \centering
  \begin{tabularx}{\textwidth}{l   s s s   s s s   s s s   s s s}
    \toprule
    {} & \multicolumn{3}{c}{AP @ IoU} & \multicolumn{3}{c}{Recall  @ IoU} & \multicolumn{3}{c}{AP @ $\Delta$DTC} & \multicolumn{3}{c}{Recall @ $\Delta$DTC} \\
    \cmidrule(lr){2-4} \cmidrule(lr){5-7} \cmidrule(lr){8-10} \cmidrule(lr){11-13}
    Supervision & 0.3 & 0.5 & 0.7 & 0.3 & 0.5 & 0.7      & 1.5 & 1.0 & 0.5 & 1.5 & 1.0 & 0.5  \\
    \midrule
    1 log (80 frames)  & 36.8 & 30.7 & 18.9 & 46.9 & 36.4 & 23.2 & 48.6 & 45.9 & 37.9 & 65.1 & 61.4 & 52.5 \\
    2 logs (160 frames)  & 51.9 & 45.0 & 29.8 & 61.9 & 52.0 & 35.0 & 56.9 & 54.5 & 46.5 & 69.9 & 67.0 & 59.3 \\
    80 random frames  & 70.6 & 61.0 & 41.1 & 80.1 & 67.6 & 46.1 & 75.6 & 73.3 & 65.8 & 87.3 & 85.0 & 78.1  \\
    160 random frames  & 81.3 & 74.2 & 55.4 & 87.1 & 78.6 & 59.7 & 84.3 & 83.1 & 77.1 & 91.3 & 90.0 & 84.8 \\
    \midrule
    DBSCAN \cite{dbscan} (unsupervised)  & 3.5 & 1.1 & 0.3 & 28.6 & 15.9 & 8.3 & 10.9 & 9.9 & 8.0 & 50.9 & 48.3 & 43.1  \\
    MODEST \cite{detect-mobile-objects} (unsupervised)  & 22.8 & 7.5 & 2.8 & 49.7 & 28.9 & 14.9 & 38.8 & 36.4 & 30.2 & 70.2 & 66.9 & 59.0 \\
    Ours (unsupervised)           & 43.5 & 29.5 & 18.1 & 62.8 & 44.8 & 28.1 & 33.8 & 44.3 & 48.7 & 54.4 & 67.0 & 72.3  \\
    \bottomrule
  \end{tabularx}
  \caption{
    \textbf{[Pandaset] Comparison against few-shot supervised methods}. All-class evaluation on the range 0-80m. 
  }
  \label{table:supervised-all-class-pandaset}
\end{table*}

%% file: figs/quantitative_ssl_iter.tex
\begin{figure*}
    \centering
    \includegraphics[width=0.45\linewidth]{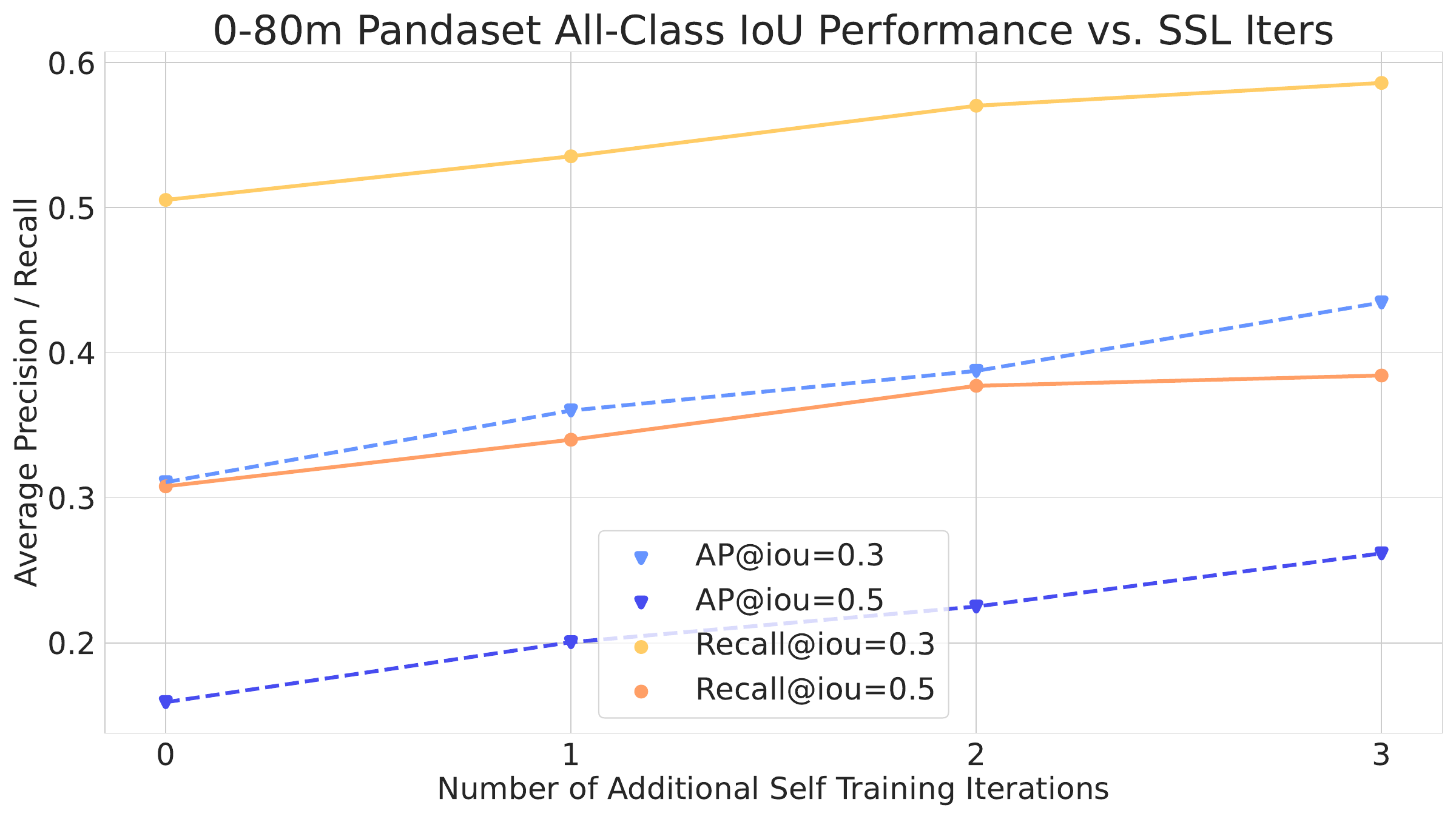}
    \includegraphics[width=0.45\linewidth]{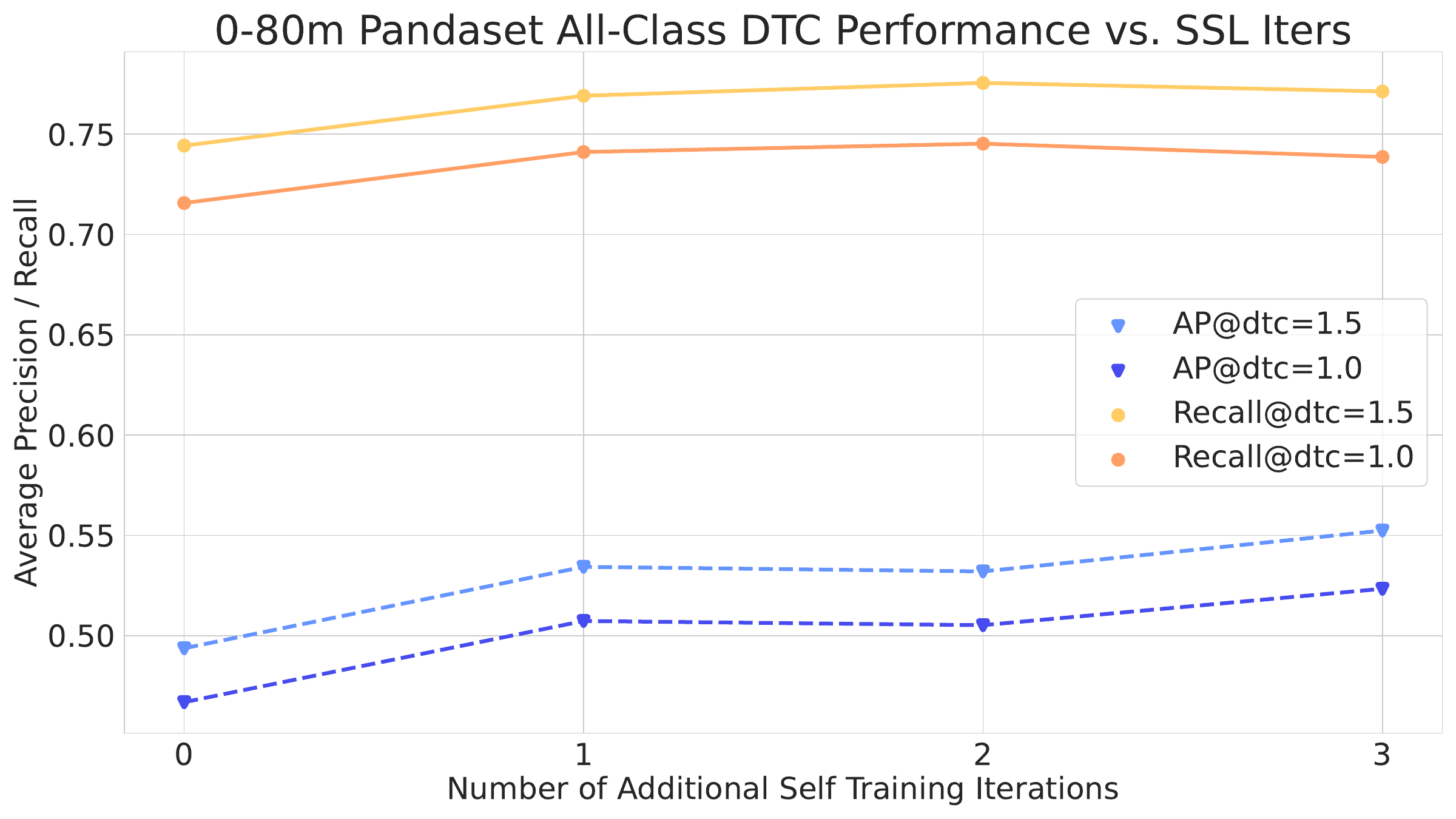}
    \includegraphics[width=0.45\linewidth]{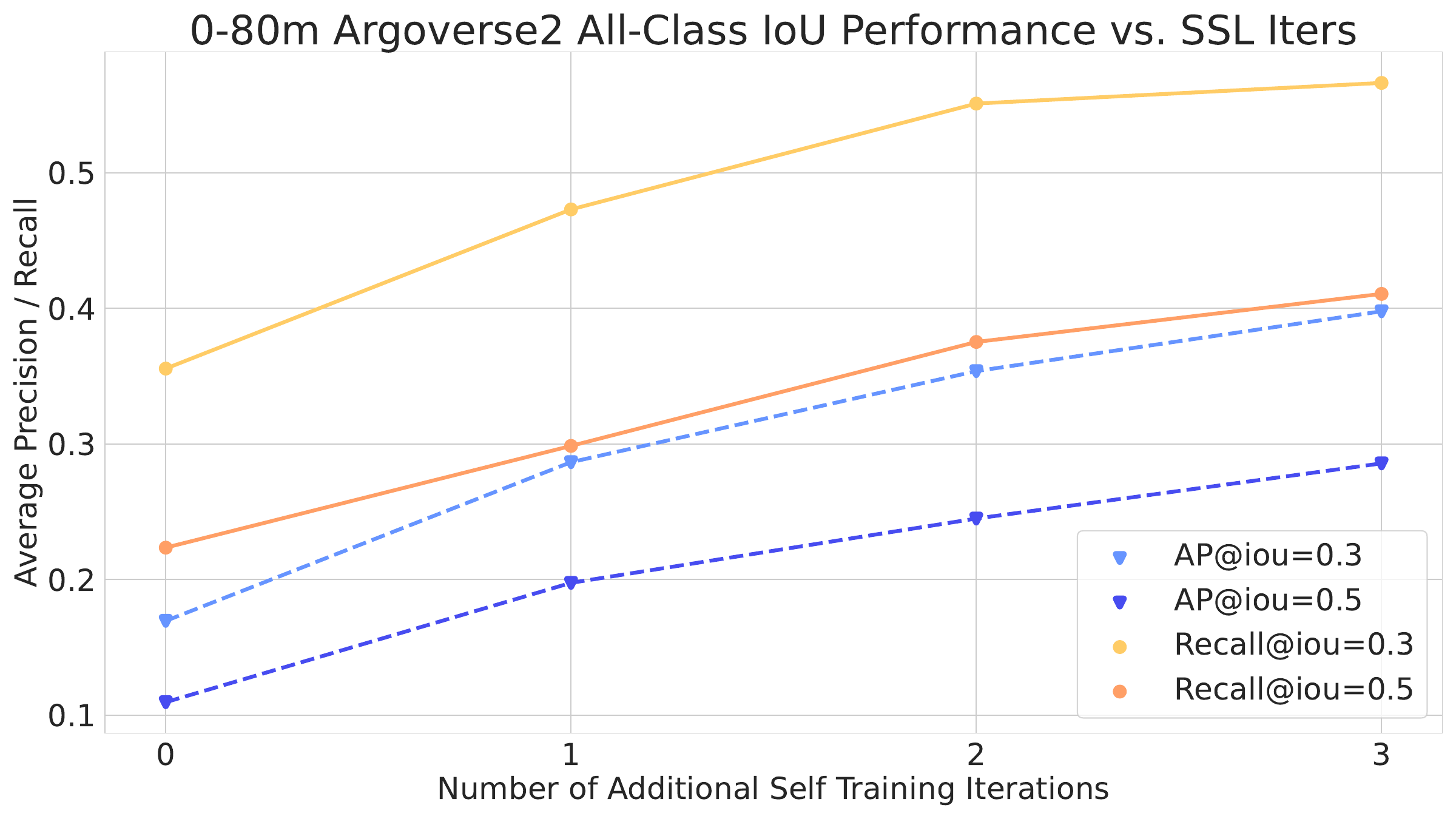}
    \includegraphics[width=0.45\linewidth]{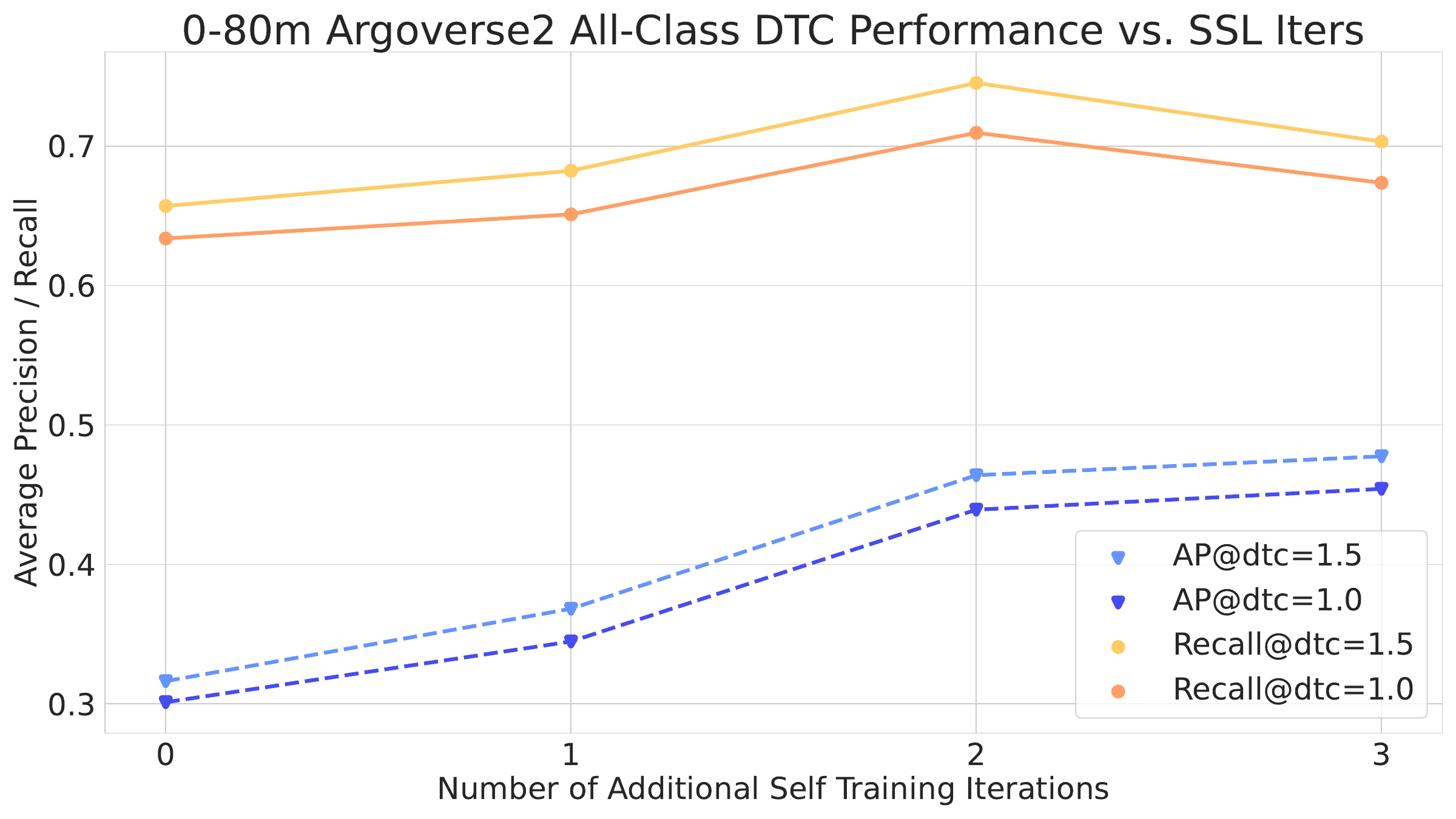}
    \caption{\textbf{Performance vs. number of self-training iterations}. We showcase the IoU (left column) and DTC (right column) mean AP and recall metrics with respect to the number of self-training iterations after the initial round. We denote the initial round results as iteration 0 in the x-axis. For both Pandaset (top row) and Argoverse (bottom row) results, more self-training iterations improved almost all metrics, while DTC-based recall dropped slightly with the third self-training iteration.}
    \label{fig:quantitative-ssl-iter}
\end{figure*}

%% file: sec/supp/qualitative_results.tex
\section{Additional Qualitative Results:}
\label{sec:supp-qual}

\input{figs/supp-qualitative-pandaset.tex}
\input{figs/supp-qualitative-argoverse.tex}
\input{figs/qualitative_refinement}

\paragraph{Qualitative results of ours vs. baselines:} Fig.~\ref{fig:supp-qualitative-comparison-pandaset} and Fig.~\ref{fig:supp-qualitative-comparison-argoverse} show additional qualitative results comparing our method and several baselines~\cite{dbscan,detect-mobile-objects} on Pandaset and AV2. Our method is able to identify stationary actors and output detections with improved sizes and positions. Please also refer to the supplementary video for more results.
\vspace{-10pt}
\paragraph{Qualitative results of refinement with self-training:} Fig.~\ref{fig:qualitative-refinement} showcases the qualitative pseudo-labels after one iteration of self-training on refined labels. We show the model output after the second iteration of self-training on Pandaset, the long tracklet refinement results on the pseudo-labels, followed by short tracklet filtering and a self-training iteration with the refined pseudo-labels. Note that the long tracklet refinement step effectively improves the bounding box sizes for some actors, and the short tracklet filtering is able to identify a few false positives and ignores them during the re-training. With an additional round of self-training, more labels are discovered and the bounding box sizes are generally improved.
\vspace{-10pt}
\paragraph{Failure cases:} 
We additionally point to various failure cases in different stages of our pipeline, including seed labels, temporal-based filtering and iterative self-training. First of all, our initial seed labels (produced by our baseline DBSCAN) have many failure cases showcased in Fig.~6 of the main paper and Fig.~2 and Fig.~3 of the supplementary. 
The DBSCAN labels in the left column of the figures contain false positives and missed detections. Note that thanks to our self-training iterations, early-stage missed detections are discovered, as shown in Fig.~6 in the main paper. 
However, there is a limit to what self-training can do. In the second row of Fig.~5 of the main paper, our method still misses detections from DBSCAN in the top left corner, even though it discovers many objects on the right side of the image. 
As for temporal consistency, in the third row of Fig.~4 in the supplementary, temporal-consistency based short tracklet filtering incorrectly filters out some cars in the parking lot.

%% file: figs/supp-qualitative-pandaset.tex
\begin{figure*}
    \vspace{-15pt}
    \centering
    \includegraphics[width=0.95\linewidth]{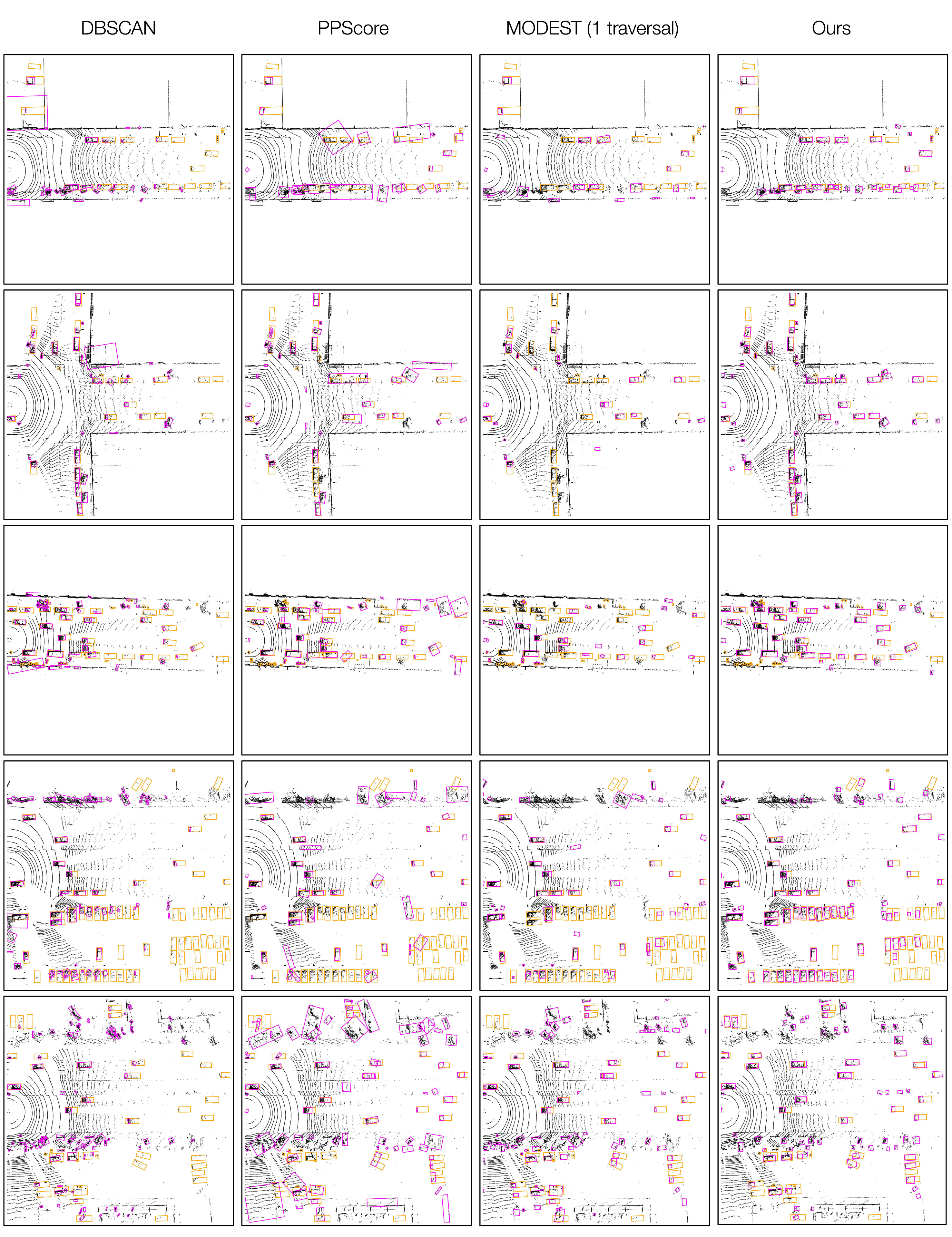}
    \vspace{-10pt}
    \caption{
        \textbf{[Pandaset] Additional qualitative results} comparing ours with baselines on frames in the validation split. We draw the detections in purple and ground-truth bounding boxes in orange.
    }
    \label{fig:supp-qualitative-comparison-pandaset}
    \vspace{-15pt}
  \end{figure*}

%% file: figs/supp-qualitative-argoverse.tex
\begin{figure*}
    \vspace{-15pt}
    \centering
    \includegraphics[width=0.95\linewidth]{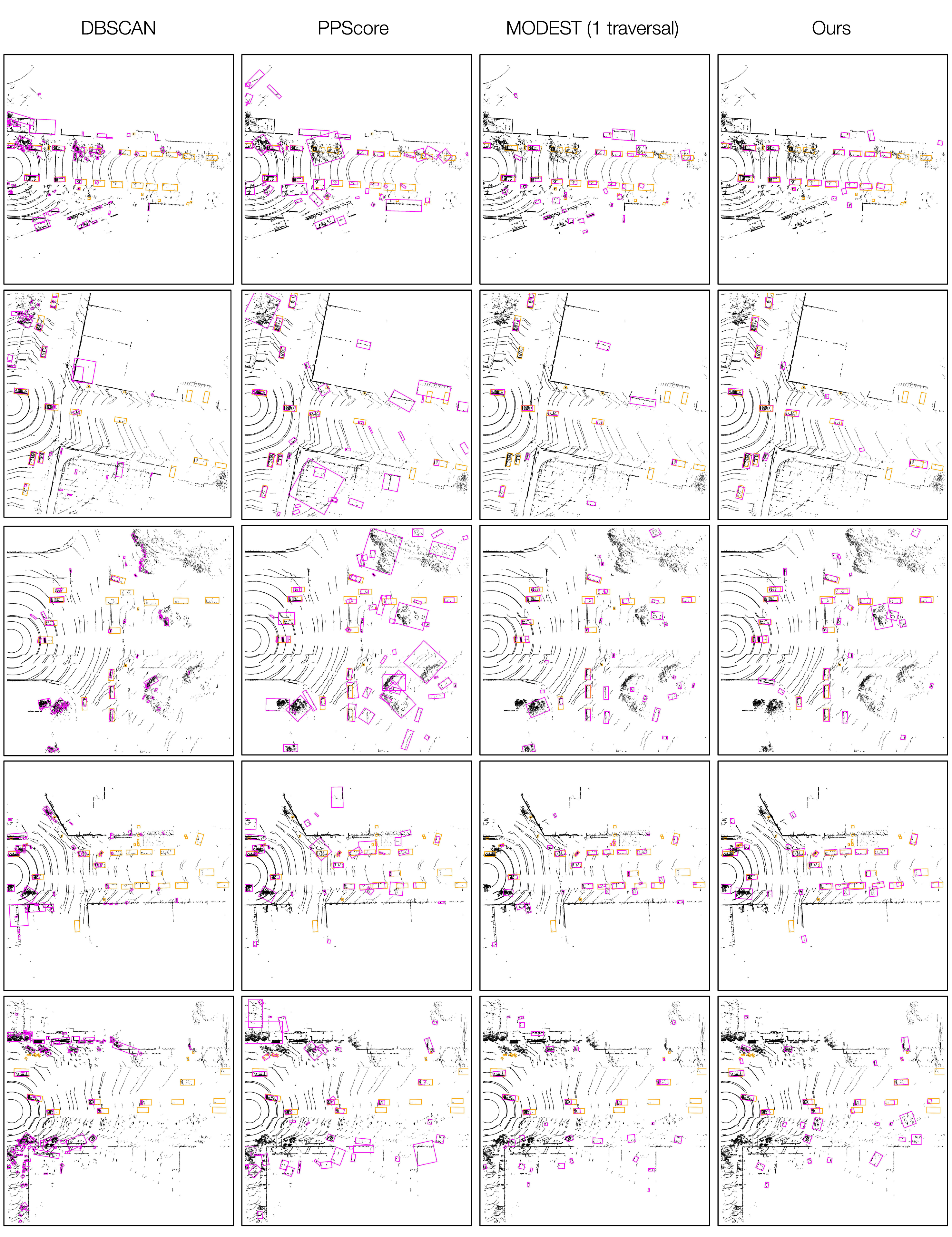}
    \vspace{-10pt}
    \caption{
        \textbf{[AV2] Additional qualitative results} comparing ours with baselines on frames in the validation split. We draw the detections in purple and ground-truth bounding boxes in orange.
    }
    \label{fig:supp-qualitative-comparison-argoverse}
    \vspace{-15pt}
  \end{figure*}

%% file: figs/qualitative_refinement.tex
\begin{figure*}
    \centering
    \includegraphics[width=0.98\linewidth]{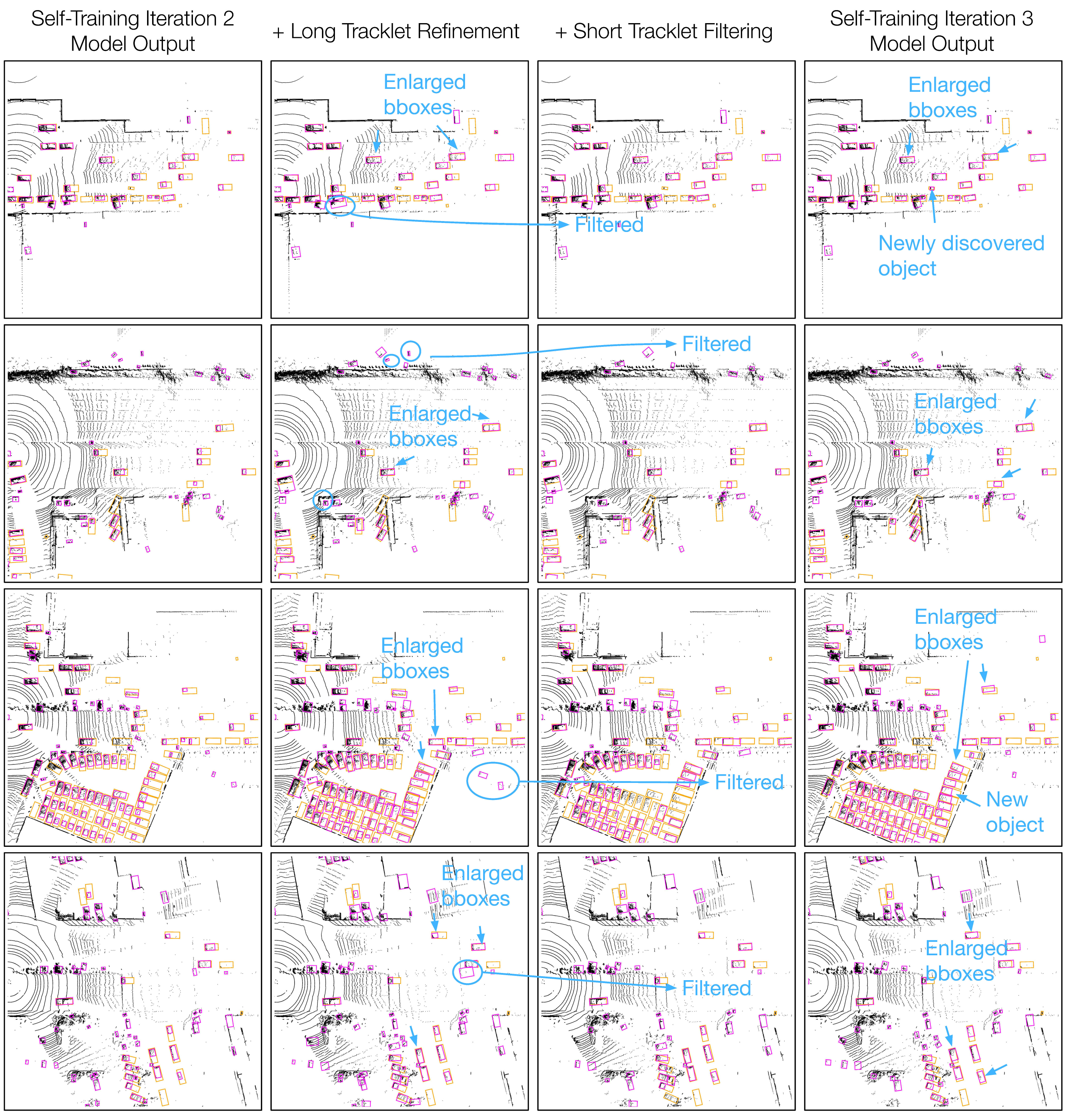}
    \caption{\textbf{[Pandaset] Qualitative results before and after self-training with refinement}. We take the model output from self-training iteration 2 on the left, apply long tracklet refinement (middle left) and short tracklet filtering (middle right), and re-train with the refined pseudo-labels, leading to new model output on the right. The new detections have larger bbox sizes and discover new objects as a result of the long tracklet refinement and self-training. We draw the detections in purple and ground-truth bounding boxes in orange.}
    \label{fig:qualitative-refinement}
\end{figure*}

%% file: PaperForReview.bbl
\begin{thebibliography}{10}\itemsep=-1pt

\bibitem{measure-objectness}
Bogdan Alexe, Thomas Deselaers, and Vittorio Ferrari.
\newblock Measuring the objectness of image windows.
\newblock {\em IEEE transactions on pattern analysis and machine intelligence}, 34(11):2189--2202, 2012.

\bibitem{deep-learning-for-ai}
Yoshua Bengio, Yann Lecun, and Geoffrey Hinton.
\newblock Deep learning for ai.
\newblock {\em Communications of the ACM}, 64(7):58--65, 2021.

\bibitem{mckayicp92}
P.J. Besl and Neil~D. McKay.
\newblock A method for registration of 3-d shapes.
\newblock {\em IEEE Transactions on Pattern Analysis and Machine Intelligence}, 14(2):239--256, 1992.

\bibitem{convex-clustering}
Hakan Bilen, Marco Pedersoli, and Tinne Tuytelaars.
\newblock Weakly supervised object detection with convex clustering.
\newblock In {\em Proceedings of the IEEE conference on computer vision and pattern recognition}, pages 1081--1089, 2015.

\bibitem{bogoslavskyi2016}
Igor Bogoslavskyi and Cyrill Stachniss.
\newblock Fast range image-based segmentation of sparse 3d laser scans for online operation.
\newblock In {\em 2016 {IEEE/RSJ} International Conference on Intelligent Robots and Systems, {IROS} 2016, Daejeon, South Korea, October 9-14, 2016}, pages 163--169. {IEEE}, 2016.

\bibitem{burgess2019monet}
Christopher~P. Burgess, Lo{\"{\i}}c Matthey, Nicholas Watters, Rishabh Kabra, Irina Higgins, Matthew Botvinick, and Alexander Lerchner.
\newblock Monet: Unsupervised scene decomposition and representation.
\newblock {\em CoRR}, abs/1901.11390, 2019.

\bibitem{detr}
Nicolas Carion, Francisco Massa, Gabriel Synnaeve, Nicolas Usunier, Alexander Kirillov, and Sergey Zagoruyko.
\newblock End-to-end object detection with transformers.
\newblock In {\em European conference on computer vision}, pages 213--229. Springer, 2020.

\bibitem{carreirasminchisescu2010}
Jo{\~{a}}o Carreira and Cristian Sminchisescu.
\newblock Constrained parametric min-cuts for automatic object segmentation.
\newblock In {\em The Twenty-Third {IEEE} Conference on Computer Vision and Pattern Recognition, {CVPR} 2010, San Francisco, CA, USA, 13-18 June 2010}, pages 3241--3248. {IEEE} Computer Society, 2010.

\bibitem{cen2021openset}
Jun Cen, Peng Yun, Junhao Cai, Michael~Yu Wang, and Ming Liu.
\newblock Open-set 3d object detection.
\newblock {\em CoRR}, abs/2112.01135, 2021.

\bibitem{crawfordpineau2019spair}
Eric Crawford and Joelle Pineau.
\newblock Spatially invariant unsupervised object detection with convolutional neural networks.
\newblock In {\em The Thirty-Third {AAAI} Conference on Artificial Intelligence, {AAAI} 2019, The Thirty-First Innovative Applications of Artificial Intelligence Conference, {IAAI} 2019, The Ninth {AAAI} Symposium on Educational Advances in Artificial Intelligence, {EAAI} 2019, Honolulu, Hawaii, USA, January 27 - February 1, 2019}, pages 3412--3420. {AAAI} Press, 2019.

\bibitem{crawford2020spairtracking}
Eric Crawford and Joelle Pineau.
\newblock Exploiting spatial invariance for scalable unsupervised object tracking.
\newblock In {\em The Thirty-Fourth {AAAI} Conference on Artificial Intelligence, {AAAI} 2020, The Thirty-Second Innovative Applications of Artificial Intelligence Conference, {IAAI} 2020, The Tenth {AAAI} Symposium on Educational Advances in Artificial Intelligence, {EAAI} 2020, New York, NY, USA, February 7-12, 2020}, pages 3684--3692. {AAAI} Press, 2020.

\bibitem{dhamija2020overlook}
Akshay~Raj Dhamija, Manuel G{\"{u}}nther, Jonathan Ventura, and Terrance~E. Boult.
\newblock The overlooked elephant of object detection: Open set.
\newblock In {\em {IEEE} Winter Conference on Applications of Computer Vision, {WACV} 2020, Snowmass Village, CO, USA, March 1-5, 2020}, pages 1010--1019. {IEEE}, 2020.

\bibitem{douillard2011segmentation}
Bertrand Douillard, James Underwood, Noah Kuntz, Vsevolod Vlaskine, Alastair Quadros, Peter Morton, and Alon Frenkel.
\newblock On the segmentation of 3d lidar point clouds.
\newblock In {\em 2011 IEEE International Conference on Robotics and Automation}, pages 2798--2805. IEEE, 2011.

\bibitem{douillard2011}
Bertrand Douillard, James~Patrick Underwood, Noah Kuntz, Vsevolod Vlaskine, Alastair~James Quadros, Peter Morton, and Alon Frenkel.
\newblock On the segmentation of 3d {LIDAR} point clouds.
\newblock In {\em {IEEE} International Conference on Robotics and Automation, {ICRA} 2011, Shanghai, China, 9-13 May 2011}, pages 2798--2805. {IEEE}, 2011.

\bibitem{engelcke2020genesis}
Martin Engelcke, Adam~R. Kosiorek, Oiwi~Parker Jones, and Ingmar Posner.
\newblock {GENESIS:} generative scene inference and sampling with object-centric latent representations.
\newblock In {\em 8th International Conference on Learning Representations, {ICLR} 2020, Addis Ababa, Ethiopia, April 26-30, 2020}. OpenReview.net, 2020.

\bibitem{eslami2016air}
S.~M.~Ali Eslami, Nicolas Heess, Theophane Weber, Yuval Tassa, David Szepesvari, Koray Kavukcuoglu, and Geoffrey~E. Hinton.
\newblock Attend, infer, repeat: Fast scene understanding with generative models.
\newblock In Daniel~D. Lee, Masashi Sugiyama, Ulrike von Luxburg, Isabelle Guyon, and Roman Garnett, editors, {\em Advances in Neural Information Processing Systems 29: Annual Conference on Neural Information Processing Systems 2016, December 5-10, 2016, Barcelona, Spain}, pages 3225--3233, 2016.

\bibitem{dbscan}
Martin Ester, Hans-Peter Kriegel, J{\"o}rg Sander, Xiaowei Xu, et~al.
\newblock A density-based algorithm for discovering clusters in large spatial databases with noise.
\newblock In {\em kdd}, volume~96, pages 226--231, 1996.

\bibitem{graph-based-segmentation}
Pedro~F Felzenszwalb and Daniel~P Huttenlocher.
\newblock Efficient graph-based image segmentation.
\newblock {\em International journal of computer vision}, 59(2):167--181, 2004.

\bibitem{RANSAC}
Martin~A Fischler and Robert~C Bolles.
\newblock Random sample consensus: a paradigm for model fitting with applications to image analysis and automated cartography.
\newblock {\em Communications of the ACM}, 24(6):381--395, 1981.

\bibitem{fragkiadaki2015}
Katerina Fragkiadaki, Pablo Arbelaez, Panna Felsen, and Jitendra Malik.
\newblock Learning to segment moving objects in videos.
\newblock In {\em {IEEE} Conference on Computer Vision and Pattern Recognition, {CVPR} 2015, Boston, MA, USA, June 7-12, 2015}, pages 4083--4090. {IEEE} Computer Society, 2015.

\bibitem{rcnn}
Ross Girshick, Jeff Donahue, Trevor Darrell, and Jitendra Malik.
\newblock Rich feature hierarchies for accurate object detection and semantic segmentation.
\newblock In {\em Proceedings of the IEEE conference on computer vision and pattern recognition}, pages 580--587, 2014.

\bibitem{greff2019iodine}
Klaus Greff, Rapha{\"{e}}l~Lopez Kaufman, Rishabh Kabra, Nick Watters, Christopher Burgess, Daniel Zoran, Loic Matthey, Matthew Botvinick, and Alexander Lerchner.
\newblock Multi-object representation learning with iterative variational inference.
\newblock In Kamalika Chaudhuri and Ruslan Salakhutdinov, editors, {\em Proceedings of the 36th International Conference on Machine Learning, {ICML} 2019, 9-15 June 2019, Long Beach, California, {USA}}, volume~97 of {\em Proceedings of Machine Learning Research}, pages 2424--2433. {PMLR}, 2019.

\bibitem{greff2017neuralem}
Klaus Greff, Sjoerd van Steenkiste, and J{\"{u}}rgen Schmidhuber.
\newblock Neural expectation maximization.
\newblock In {\em Advances in Neural Information Processing Systems 30: Annual Conference on Neural Information Processing Systems 2017, December 4-9, 2017, Long Beach, CA, {USA}}, pages 6691--6701, 2017.

\bibitem{gupta2021owdetr}
Akshita Gupta, Sanath Narayan, K.~J. Joseph, Salman~H. Khan, Fahad~Shahbaz Khan, and Mubarak Shah.
\newblock {OW-DETR:} open-world detection transformer.
\newblock {\em CoRR}, abs/2112.01513, 2021.

\bibitem{mask-rcnn}
Kaiming He, Georgia Gkioxari, Piotr Doll{\'a}r, and Ross Girshick.
\newblock Mask r-cnn.
\newblock In {\em Proceedings of the IEEE international conference on computer vision}, pages 2961--2969, 2017.

\bibitem{resnet}
Kaiming He, Xiangyu Zhang, Shaoqing Ren, and Jian Sun.
\newblock Deep residual learning for image recognition.
\newblock In {\em Proceedings of the IEEE conference on computer vision and pattern recognition}, pages 770--778, 2016.

\bibitem{himmelsbach2010fast}
Michael Himmelsbach, Felix~V Hundelshausen, and H-J Wuensche.
\newblock Fast segmentation of 3d point clouds for ground vehicles.
\newblock In {\em 2010 IEEE Intelligent Vehicles Symposium}, pages 560--565. IEEE, 2010.

\bibitem{batchnorm}
Sergey Ioffe and Christian Szegedy.
\newblock Batch normalization: Accelerating deep network training by reducing internal covariate shift.
\newblock In {\em International conference on machine learning}, pages 448--456. PMLR, 2015.

\bibitem{jaques2020physics}
Miguel Jaques, Michael Burke, and Timothy~M. Hospedales.
\newblock Physics-as-inverse-graphics: Unsupervised physical parameter estimation from video.
\newblock In {\em 8th International Conference on Learning Representations, {ICLR} 2020, Addis Ababa, Ethiopia, April 26-30, 2020}. OpenReview.net, 2020.

\bibitem{jiang2020generative}
Jindong Jiang and Sungjin Ahn.
\newblock Generative neurosymbolic machines.
\newblock In Hugo Larochelle, Marc'Aurelio Ranzato, Raia Hadsell, Maria{-}Florina Balcan, and Hsuan{-}Tien Lin, editors, {\em Advances in Neural Information Processing Systems 33: Annual Conference on Neural Information Processing Systems 2020, NeurIPS 2020, December 6-12, 2020, virtual}, 2020.

\bibitem{jiang2020scalor}
Jindong Jiang, Sepehr Janghorbani, Gerard de Melo, and Sungjin Ahn.
\newblock {SCALOR:} generative world models with scalable object representations.
\newblock In {\em 8th International Conference on Learning Representations, {ICLR} 2020, Addis Ababa, Ethiopia, April 26-30, 2020}. OpenReview.net, 2020.

\bibitem{joseph2021openworld}
K.~J. Joseph, Salman~H. Khan, Fahad~Shahbaz Khan, and Vineeth~N. Balasubramanian.
\newblock Towards open world object detection.
\newblock In {\em {IEEE} Conference on Computer Vision and Pattern Recognition, {CVPR} 2021, virtual, June 19-25, 2021}, pages 5830--5840. Computer Vision Foundation / {IEEE}, 2021.

\bibitem{kabra2021simone}
Rishabh Kabra, Daniel Zoran, Goker Erdogan, Loic Matthey, Antonia Creswell, Matthew Botvinick, Alexander Lerchner, and Christopher~P. Burgess.
\newblock Simone: View-invariant, temporally-abstracted object representations via unsupervised video decomposition.
\newblock {\em CoRR}, abs/2106.03849, 2021.

\bibitem{karpathy2013}
Andrej Karpathy, Stephen~D. Miller, and Li Fei{-}Fei.
\newblock Object discovery in 3d scenes via shape analysis.
\newblock In {\em 2013 {IEEE} International Conference on Robotics and Automation, Karlsruhe, Germany, May 6-10, 2013}, pages 2088--2095. {IEEE}, 2013.

\bibitem{adam-optimizer}
Diederik~P Kingma and Jimmy Ba.
\newblock Adam: A method for stochastic optimization.
\newblock {\em arXiv preprint arXiv:1412.6980}, 2014.

\bibitem{kosiorek2018sequentialair}
Adam~R. Kosiorek, Hyunjik Kim, Yee~Whye Teh, and Ingmar Posner.
\newblock Sequential attend, infer, repeat: Generative modelling of moving objects.
\newblock In Samy Bengio, Hanna~M. Wallach, Hugo Larochelle, Kristen Grauman, Nicol{\`{o}} Cesa{-}Bianchi, and Roman Garnett, editors, {\em Advances in Neural Information Processing Systems 31: Annual Conference on Neural Information Processing Systems 2018, NeurIPS 2018, December 3-8, 2018, Montr{\'{e}}al, Canada}, pages 8615--8625, 2018.

\bibitem{kosiorek2019scae}
Adam~R. Kosiorek, Sara Sabour, Yee~Whye Teh, and Geoffrey~E. Hinton.
\newblock Stacked capsule autoencoders.
\newblock In Hanna~M. Wallach, Hugo Larochelle, Alina Beygelzimer, Florence d'Alch{\'{e}}{-}Buc, Emily~B. Fox, and Roman Garnett, editors, {\em Advances in Neural Information Processing Systems 32: Annual Conference on Neural Information Processing Systems 2019, NeurIPS 2019, December 8-14, 2019, Vancouver, BC, Canada}, pages 15486--15496, 2019.

\bibitem{pointpillars}
Alex~H Lang, Sourabh Vora, Holger Caesar, Lubing Zhou, Jiong Yang, and Oscar Beijbom.
\newblock Pointpillars: Fast encoders for object detection from point clouds.
\newblock In {\em Proceedings of the IEEE/CVF conference on computer vision and pattern recognition}, pages 12697--12705, 2019.

\bibitem{fpn-net}
Tsung-Yi Lin, Piotr Doll{\'a}r, Ross Girshick, Kaiming He, Bharath Hariharan, and Serge Belongie.
\newblock Feature pyramid networks for object detection.
\newblock In {\em Proceedings of the IEEE conference on computer vision and pattern recognition}, pages 2117--2125, 2017.

\bibitem{fpn}
Tsung-Yi Lin, Piotr Doll{\'a}r, Ross Girshick, Kaiming He, Bharath Hariharan, and Serge Belongie.
\newblock Feature pyramid networks for object detection.
\newblock In {\em Proceedings of the IEEE conference on computer vision and pattern recognition}, pages 2117--2125, 2017.

\bibitem{focal-loss}
Tsung-Yi Lin, Priya Goyal, Ross Girshick, Kaiming He, and Piotr Doll{\'a}r.
\newblock Focal loss for dense object detection.
\newblock In {\em Proceedings of the IEEE international conference on computer vision}, pages 2980--2988, 2017.

\bibitem{lin2020space}
Zhixuan Lin, Yi{-}Fu Wu, Skand~Vishwanath Peri, Weihao Sun, Gautam Singh, Fei Deng, Jindong Jiang, and Sungjin Ahn.
\newblock {SPACE:} unsupervised object-oriented scene representation via spatial attention and decomposition.
\newblock In {\em 8th International Conference on Learning Representations, {ICLR} 2020, Addis Ababa, Ethiopia, April 26-30, 2020}. OpenReview.net, 2020.

\bibitem{locatello2020slot}
Francesco Locatello, Dirk Weissenborn, Thomas Unterthiner, Aravindh Mahendran, Georg Heigold, Jakob Uszkoreit, Alexey Dosovitskiy, and Thomas Kipf.
\newblock Object-centric learning with slot attention.
\newblock In Hugo Larochelle, Marc'Aurelio Ranzato, Raia Hadsell, Maria{-}Florina Balcan, and Hsuan{-}Tien Lin, editors, {\em Advances in Neural Information Processing Systems 33: Annual Conference on Neural Information Processing Systems 2020, NeurIPS 2020, December 6-12, 2020, virtual}, 2020.

\bibitem{adamw}
Ilya Loshchilov and Frank Hutter.
\newblock Decoupled weight decay regularization.
\newblock {\em arXiv preprint arXiv:1711.05101}, 2017.

\bibitem{mildenhall2020nerf}
Ben Mildenhall, Pratul~P. Srinivasan, Matthew Tancik, Jonathan~T. Barron, Ravi Ramamoorthi, and Ren Ng.
\newblock Nerf: Representing scenes as neural radiance fields for view synthesis.
\newblock In Andrea Vedaldi, Horst Bischof, Thomas Brox, and Jan{-}Michael Frahm, editors, {\em Computer Vision - {ECCV} 2020 - 16th European Conference, Glasgow, UK, August 23-28, 2020, Proceedings, Part {I}}, volume 12346 of {\em Lecture Notes in Computer Science}, pages 405--421. Springer, 2020.

\bibitem{miller2018dropout}
Dimity Miller, Lachlan Nicholson, Feras Dayoub, and Niko S{\"{u}}nderhauf.
\newblock Dropout sampling for robust object detection in open-set conditions.
\newblock In {\em 2018 {IEEE} International Conference on Robotics and Automation, {ICRA} 2018, Brisbane, Australia, May 21-25, 2018}, pages 1--7. {IEEE}, 2018.

\bibitem{najibi2022motion}
Mahyar Najibi, Jingwei Ji, Yin Zhou, Charles~R Qi, Xinchen Yan, Scott Ettinger, and Dragomir Anguelov.
\newblock Motion inspired unsupervised perception and prediction in autonomous driving.
\newblock In {\em European Conference on Computer Vision}, pages 424--443. Springer, 2022.

\bibitem{namdev2012}
Rahul~Kumar Namdev, Abhijit Kundu, K.~Madhava Krishna, and C.~V. Jawahar.
\newblock Motion segmentation of multiple objects from a freely moving monocular camera.
\newblock In {\em {IEEE} International Conference on Robotics and Automation, {ICRA} 2012, 14-18 May, 2012, St. Paul, Minnesota, {USA}}, pages 4092--4099. {IEEE}, 2012.

\bibitem{papazoglouferrari2013}
Anestis Papazoglou and Vittorio Ferrari.
\newblock Fast object segmentation in unconstrained video.
\newblock In {\em {IEEE} International Conference on Computer Vision, {ICCV} 2013, Sydney, Australia, December 1-8, 2013}, pages 1777--1784. {IEEE} Computer Society, 2013.

\bibitem{pham2018bayesian}
Trung Pham, B.~G.~Vijay Kumar, Thanh{-}Toan Do, Gustavo Carneiro, and Ian~D. Reid.
\newblock Bayesian semantic instance segmentation in open set world.
\newblock In Vittorio Ferrari, Martial Hebert, Cristian Sminchisescu, and Yair Weiss, editors, {\em Computer Vision - {ECCV} 2018 - 15th European Conference, Munich, Germany, September 8-14, 2018, Proceedings, Part {X}}, volume 11214 of {\em Lecture Notes in Computer Science}, pages 3--18. Springer, 2018.

\bibitem{ponttuset2017}
Jordi Pont{-}Tuset, Pablo Arbelaez, Jonathan~T. Barron, Ferran Marqu{\'{e}}s, and Jitendra Malik.
\newblock Multiscale combinatorial grouping for image segmentation and object proposal generation.
\newblock {\em {IEEE} Trans. Pattern Anal. Mach. Intell.}, 39(1):128--140, 2017.

\bibitem{faster-rcnn}
Shaoqing Ren, Kaiming He, Ross Girshick, and Jian Sun.
\newblock Faster r-cnn: Towards real-time object detection with region proposal networks.
\newblock {\em Advances in neural information processing systems}, 28, 2015.

\bibitem{sabour2021flowcapsule}
Sara Sabour, Andrea Tagliasacchi, Soroosh Yazdani, Geoffrey~E. Hinton, and David~J. Fleet.
\newblock Unsupervised part representation by flow capsules.
\newblock In Marina Meila and Tong Zhang, editors, {\em Proceedings of the 38th International Conference on Machine Learning, {ICML} 2021, 18-24 July 2021, Virtual Event}, volume 139 of {\em Proceedings of Machine Learning Research}, pages 9213--9223. {PMLR}, 2021.

\bibitem{sivic2005}
Josef Sivic, Bryan~C. Russell, Alexei~A. Efros, Andrew Zisserman, and William~T. Freeman.
\newblock Discovering objects and their localization in images.
\newblock In {\em 10th {IEEE} International Conference on Computer Vision {(ICCV} 2005), 17-20 October 2005, Beijing, China}, pages 370--377. {IEEE} Computer Society, 2005.

\bibitem{motion-cue-detect}
Andrew Stein, Derek Hoiem, and Martial Hebert.
\newblock Learning to find object boundaries using motion cues.
\newblock In {\em 2007 IEEE 11th International Conference on Computer Vision}, pages 1--8. IEEE, 2007.

\bibitem{stelzner2021obsurf}
Karl Stelzner, Kristian Kersting, and Adam~R. Kosiorek.
\newblock Decomposing 3d scenes into objects via unsupervised volume segmentation.
\newblock {\em CoRR}, abs/2104.01148, 2021.

\bibitem{uijlings2013}
Jasper R.~R. Uijlings, Koen E.~A. van~de Sande, Theo Gevers, and Arnold W.~M. Smeulders.
\newblock Selective search for object recognition.
\newblock {\em Int. J. Comput. Vis.}, 104(2):154--171, 2013.

\bibitem{steenkiste2018relationalnem}
Sjoerd van Steenkiste, Michael Chang, Klaus Greff, and J{\"{u}}rgen Schmidhuber.
\newblock Relational neural expectation maximization: Unsupervised discovery of objects and their interactions.
\newblock In {\em 6th International Conference on Learning Representations, {ICLR} 2018, Vancouver, BC, Canada, April 30 - May 3, 2018, Conference Track Proceedings}. OpenReview.net, 2018.

\bibitem{vicente2011coseg}
Sara Vicente, Carsten Rother, and Vladimir Kolmogorov.
\newblock Object cosegmentation.
\newblock In {\em The 24th {IEEE} Conference on Computer Vision and Pattern Recognition, {CVPR} 2011, Colorado Springs, CO, USA, 20-25 June 2011}, pages 2217--2224. {IEEE} Computer Society, 2011.

\bibitem{vo2020discovery}
Huy~V. Vo, Patrick P{\'{e}}rez, and Jean Ponce.
\newblock Toward unsupervised, multi-object discovery in large-scale image collections.
\newblock In Andrea Vedaldi, Horst Bischof, Thomas Brox, and Jan{-}Michael Frahm, editors, {\em Computer Vision - {ECCV} 2020 - 16th European Conference, Glasgow, UK, August 23-28, 2020, Proceedings, Part {XXIII}}, volume 12368 of {\em Lecture Notes in Computer Science}, pages 779--795. Springer, 2020.

\bibitem{vo2021discovery}
Huy~V. Vo, Elena Sizikova, Cordelia Schmid, Patrick P{\'{e}}rez, and Jean Ponce.
\newblock Large-scale unsupervised object discovery.
\newblock {\em CoRR}, abs/2106.06650, 2021.

\bibitem{min-entropy-weakly-supervised}
Fang Wan, Pengxu Wei, Jianbin Jiao, Zhenjun Han, and Qixiang Ye.
\newblock Min-entropy latent model for weakly supervised object detection.
\newblock In {\em Proceedings of the IEEE Conference on Computer Vision and Pattern Recognition}, pages 1297--1306, 2018.

\bibitem{wang2021spair3d}
Tianyu Wang, Miaomiao Liu, and Kee~Siong Ng.
\newblock Spatially invariant unsupervised 3d object segmentation with graph neural networks.
\newblock {\em CoRR}, abs/2106.05607, 2021.

\bibitem{4d-unsupervised-object-discovery}
Yuqi Wang, Yuntao Chen, and Zhaoxiang Zhang.
\newblock 4d unsupervised object discovery.
\newblock {\em arXiv preprint arXiv:2210.04801}, 2022.

\bibitem{wilson2021argoverse}
Benjamin Wilson, William Qi, Tanmay Agarwal, John Lambert, Jagjeet Singh, Siddhesh Khandelwal, Bowen Pan, Ratnesh Kumar, Andrew Hartnett, Jhony~Kaesemodel Pontes, et~al.
\newblock Argoverse 2: Next generation datasets for self-driving perception and forecasting.
\newblock In {\em Thirty-fifth Conference on Neural Information Processing Systems Datasets and Benchmarks Track (Round 2)}, 2021.

\bibitem{wong2019osis}
Kelvin Wong, Shenlong Wang, Mengye Ren, Ming Liang, and Raquel Urtasun.
\newblock Identifying unknown instances for autonomous driving.
\newblock In Leslie~Pack Kaelbling, Danica Kragic, and Komei Sugiura, editors, {\em 3rd Annual Conference on Robot Learning, CoRL 2019, Osaka, Japan, October 30 - November 1, 2019, Proceedings}, volume 100 of {\em Proceedings of Machine Learning Research}, pages 384--393. {PMLR}, 2019.

\bibitem{xiao2016}
Fanyi Xiao and Yong~Jae Lee.
\newblock Track and segment: An iterative unsupervised approach for video object proposals.
\newblock In {\em 2016 {IEEE} Conference on Computer Vision and Pattern Recognition, {CVPR} 2016, Las Vegas, NV, USA, June 27-30, 2016}, pages 933--942. {IEEE} Computer Society, 2016.

\bibitem{xiao2021pandaset}
Pengchuan Xiao, Zhenlei Shao, Steven Hao, Zishuo Zhang, Xiaolin Chai, Judy Jiao, Zesong Li, Jian Wu, Kai Sun, Kun Jiang, et~al.
\newblock Pandaset: Advanced sensor suite dataset for autonomous driving.
\newblock In {\em 2021 IEEE International Intelligent Transportation Systems Conference (ITSC)}, pages 3095--3101. IEEE, 2021.

\bibitem{auto4d}
Bin Yang, Min Bai, Ming Liang, Wenyuan Zeng, and Raquel Urtasun.
\newblock Auto4d: Learning to label 4d objects from sequential point clouds, 2021.

\bibitem{yang2018pixor}
Bin Yang, Wenjie Luo, and Raquel Urtasun.
\newblock Pixor: Real-time 3d object detection from point clouds.
\newblock In {\em Proceedings of the IEEE conference on Computer Vision and Pattern Recognition}, pages 7652--7660, 2018.

\bibitem{detect-mobile-objects}
Yurong You, Katie Luo, Cheng~Perng Phoo, Wei-Lun Chao, Wen Sun, Bharath Hariharan, Mark Campbell, and Kilian~Q Weinberger.
\newblock Learning to detect mobile objects from lidar scans without labels.
\newblock In {\em CVPR}, 2022.

\bibitem{yu2021objectradiance}
Hong{-}Xing Yu, Leonidas~J. Guibas, and Jiajun Wu.
\newblock Unsupervised discovery of object radiance fields.
\newblock {\em CoRR}, abs/2107.07905, 2021.

\bibitem{zigzag-learning}
Xiaopeng Zhang, Jiashi Feng, Hongkai Xiong, and Qi Tian.
\newblock Zigzag learning for weakly supervised object detection.
\newblock In {\em Proceedings of the IEEE Conference on Computer Vision and Pattern Recognition}, pages 4262--4270, 2018.

\bibitem{bbox-fitting-algo}
Xiao Zhang, Wenda Xu, Chiyu Dong, and John~M Dolan.
\newblock Efficient l-shape fitting for vehicle detection using laser scanners.
\newblock In {\em 2017 IEEE Intelligent Vehicles Symposium (IV)}, pages 54--59. IEEE, 2017.

\bibitem{rgiou-loss}
Yu Zheng, Danyang Zhang, Sinan Xie, Jiwen Lu, and Jie Zhou.
\newblock Rotation-robust intersection over union for 3d object detection.
\newblock In {\em European Conference on Computer Vision}, pages 464--480. Springer, 2020.

\bibitem{center-net}
Xingyi Zhou, Dequan Wang, and Philipp Kr{\"a}henb{\"u}hl.
\newblock Objects as points.
\newblock {\em arXiv preprint arXiv:1904.07850}, 2019.

\end{thebibliography}
